\documentclass[letterpaper,twocolumn,english,american,10pt, twocolumn, twoside]{IEEEtran}
\usepackage[T1]{fontenc}
\usepackage[latin9]{inputenc}
\usepackage{color}
\usepackage{babel}
\usepackage{array}
\usepackage{verbatim}
\usepackage{multirow}
\usepackage{amsmath}
\usepackage{graphicx}
\PassOptionsToPackage{normalem}{ulem}
\usepackage{ulem}
\usepackage[unicode=true,pdfusetitle,
 bookmarks=true,bookmarksnumbered=false,bookmarksopen=false,
 breaklinks=false,pdfborder={0 0 1},backref=false,colorlinks=true]
 {hyperref}

\makeatletter

\pdfpageheight\paperheight
\pdfpagewidth\paperwidth

\providecommand{\tabularnewline}{\\}

\let\oldforeign@language\foreign@language
\DeclareRobustCommand{\foreign@language}[1]{%
  \lowercase{\oldforeign@language{#1}}}

\usepackage[dvipsnames,table,xcdraw]{xcolor}

\usepackage{cite}
\usepackage{amsmath,amsfonts}
\usepackage{algorithmic}
\usepackage{array}
\usepackage[caption=false,font=footnotesize,labelfont=sf,textfont=sf]{subfig}
\usepackage{textcomp}
\usepackage{stfloats}
\usepackage{url}
\usepackage{verbatim}
\usepackage{graphicx}
\usepackage{balance}

\usepackage{adjustbox}
\usepackage{mathtools}
\usepackage{alphalph}

\usepackage{xspace}
\newcommand{\etal}{\emph{et al.}\xspace}
\newcommand{\ie}{\emph{i.e.}\xspace}

\newcommand{\kitti}{KITTI\xspace}
\newcommand{\tartanair}{TartanAir\xspace}
\newcommand{\midair}{Mid-Air\xspace}

\usepackage{colortbl}
\usepackage{color}

\usepackage{environ}         
\usepackage{etoolbox}        

\newlength{\myl}
\let\origequation=\equation
\let\origendequation=\endequation

\RenewEnviron{equation}{
  \settowidth{\myl}{$\BODY$}                       
  \origequation
  \ifdimcomp{0.89\linewidth}{>}{\the\myl}
  {\ensuremath{\displaystyle\BODY}}                             
  {\resizebox{0.89\linewidth}{!}{\ensuremath{\displaystyle\BODY}}}  
  \origendequation
}

\def\merge{\Vrai} 

\@ifundefined{showcaptionsetup}{}{%
 \PassOptionsToPackage{caption=false}{subfig}}
\usepackage{subfig}
\makeatother

\begin{document}
\selectlanguage{english}%



\global\long\def\imSymbol{I}%
 
\global\long\def\imWidth{W}%
 
\global\long\def\imHeight{H}%
 
\global\long\def\hImCoord{i}%
 
\global\long\def\vImCoord{j}%


\global\long\def\focal#1{f_{#1}}%


\global\long\def\transformationMatrix#1{\text{\textbf{T}}_{#1}}%

\global\long\def\rotMatrix#1{\mathbf{R}_{#1}}%

\global\long\def\transpositionSymbol{T}%

\global\long\def\translationVector{\mathbf{\mathbf{t}}}%

\global\long\def\translationVectorComp{\mathrm{t}}%

\global\long\def\time{t}%

\global\long\def\indexOfVirtualCamera{V}%


\global\long\def\dmap{\mathbf{d}}%
 
\global\long\def\dpix{\text{z}}%
 
\global\long\def\disp{\rho}%
 
\global\long\def\dispmap{\boldsymbol{\rho}}%
\global\long\def\dispmapEstimate{\boldsymbol{{\hat{\rho}}}}%

\global\long\def\firstIndex{\time-1}%

\global\long\def\secondIndex{\time}%


\global\long\def\upscaled#1{\boldsymbol{\hat{\ }}#1}%

\global\long\def\levelIndex{l}%


\global\long\def\comma{\,,}%

\global\long\def\point{\,.}%

\newcommand{\eqx}[1]{equation\ (#1)\xspace}               
\newcommand{\eqxy}[2]{equations\ (#1) and\ (#2)\xspace}   
\newcommand{\eqxyz}[3]{equations\ (#1),\ (#2), and\ (#3)\xspace}   
\newcommand{\Eqx}[1]{Equ.\ (#1)\xspace}                   
\newcommand{\Eqxy}[2]{Equ.\ (#1) and\ (#2)\xspace}        
\newcommand{\Eqxyz}[3]{Equ.\ (#1),\ (#2), and\ (#3)\xspace}        

\newcommand{\figx}[1]{Figure\ #1\xspace}                  
\newcommand{\figxy}[2]{Figures\ #1 and\ #2\xspace}        
\newcommand{\Figx}[1]{Fig.\ #1\xspace}                    
\newcommand{\Figxy}[2]{Fig.\ #1 and\ #2\xspace}           
\newcommand{\tabx}[1]{Table\ #1\xspace}                   

\newcommand{\secx}[1]{Section\ #1\xspace}                 
\newcommand{\secxy}[2]{Sections\ #1 and\ #2\xspace}       
\newcommand{\secxyzt}[4]{Sections\ #1,\ #2,\ #3, and\ #4\xspace}
\newcommand{\chapx}[1]{Chapter\ #1\xspace}                
\newcommand{\chapxy}[2]{Chapters\ #1 and #2\xspace}                

\newcommand{\midairgithub}{\url{https://github.com/michael-fonder/M4Depth}}\selectlanguage{american}%

\title{M4Depth: Monocular depth estimation for autonomous vehicles in unseen
environments}
\author{Michaël Fonder, Damien Ernst, and Marc Van Droogenbroeck\IEEEcompsocitemizethanks{\IEEEcompsocthanksitem M. Fonder, D. Ernst, and M. Van Droogenbroeck are with the Dept. of Electr. Engineering and Computer Science, University of Liège, Belgium\protect\\
E-mail: \{Michael.Fonder,dernst,M.VanDroogenbroeck\}@uliege.be}}

\maketitle
\ifx \merge\False
\markboth{IEEE Transactions on Image Processing}{}

\fi

\IEEEpeerreviewmaketitle
\begin{abstract}
Estimating the distance to objects is crucial for autonomous vehicles
when using depth sensors is not possible. In this case, the distance
has to be estimated from on-board mounted RGB cameras, which is a
complex task especially in environments such as natural outdoor landscapes.
In this paper, we present a new method named M4Depth for depth estimation.
First, we establish a bijective relationship between depth and the
visual disparity of two consecutive frames and show how to exploit
it to perform motion-invariant pixel-wise depth estimation. Then,
we detail M4Depth which is based on a pyramidal convolutional neural
network architecture where each level refines an input disparity map
estimate by using two customized cost volumes. We use these cost volumes
to leverage the visual spatio-temporal constraints imposed by motion
and to make the network robust for varied scenes. We benchmarked our
approach both in test and generalization modes on public datasets
featuring synthetic camera trajectories recorded in a wide variety
of outdoor scenes. Results show that our network outperforms the state
of the art on these datasets, while also performing well on a standard
depth estimation benchmark. The code of our method is publicly available
at \midairgithub. 
\end{abstract}


\section{Introduction \label{sec:Introduction}}

\IEEEPARstart{E}{estimating} accurate dense depth maps is an essential
task for the planning of unmanned vehicle trajectories~\cite{Achtelik2009Stereo,Dudek2010Computational}.
To the best of our knowledge, the only reliable long-range distance
sensors suitable for outdoor are heavy, bulky and power consuming.
Consequently, they can hardly be used on vehicles such as drones where
weight, size, and power availability are constrained. Distances between
objects and the camera therefore need to be inferred instead of being
measured. This is possible when an on-board RGB camera mounted on
a vehicle is led to see the same objects from multiple viewpoints.
\begin{figure}
\begin{centering}
\includegraphics[width=0.9\linewidth]{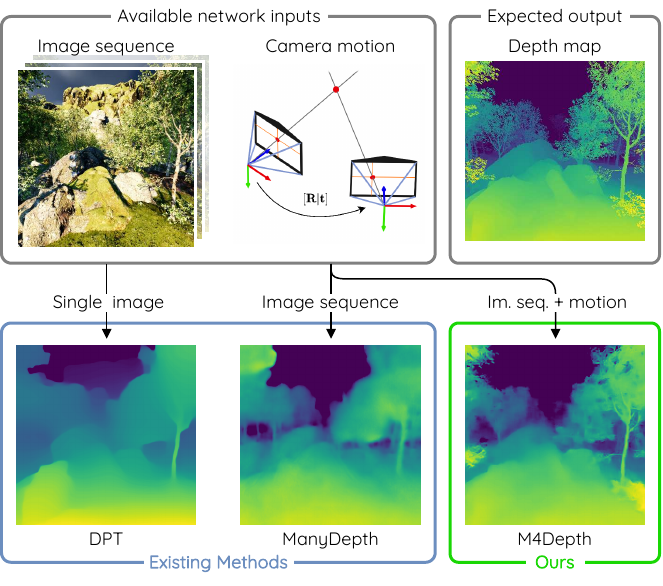}
\par\end{centering}
\caption{State-of-the-art depth estimation methods such as DPT~\cite{Ranftl2021Vision,Ranftl2022Towards}
or ManyDepth~\cite{Watson2021TheTemporal} struggle to produce accurate
estimates in cluttered and natural environments. Our method, called
M4Depth, outperforms existing methods in these instances and generalizes
well to unknown environments. \label{fig:graphical-abstract}}
\end{figure}

Small and lightweight unmanned vehicles are perfect devices to reach
places otherwise barely accessible and, hence, are often use to venture
off road in environments where direct cues about distance, such as
object structures, are unknown. Natural landscapes are examples of
such environments whose elements (vegetation, ground and relief) do
not exhibit normalized structures or patterns. Datasets suitable to
study the task of depth estimation in environments with little structure
---we call them ``unstructured environments'' in the following---
were proposed only recently (see \cite{Fonder2019MidAir,Wang2020TartanAir}).
As a result, existing methods for outdoor depth estimation are optimized
for older benchmarks targeting autonomous driving applications, on
datasets featuring constrained trajectories and environments. Indeed,
the motion of a car on a road is strongly constrained, and urban environments
contain many objects with a specific structure, such as cars, roads,
signs, buildings, etc. Since structure and semantics of elements in
a scene likely provide direct cues about depth, it is uncertain if
existing methods perform well in environments where such cues are
not available.

In this paper, we address the challenging task of estimating depth
in unstructured environments. The formulation of this task and the
related work are presented in \secxy{\ref{sec:problem-statement}}{\ref{sec:related-work}},
respectively. Then, in \secx{\ref{sec:M4Depth-method}}, we describe
a dedicated depth estimation method based on a deep neural network.
To perform well when inferring depth for specific environments or
in generalization, we use both the image sequence and the camera motion,
which enables one to derive depth from pixel displacement estimates.
In \secx{\ref{sec:Experiments}}, we detail our experimental setup
which is aimed at evaluating methods on unstructured environments
and in generalization, present results, and discuss our method. \secx{\ref{sec:Conclusion}}
concludes the paper.

Our contributions can be summarized as follows:
\begin{enumerate}
\item We extend the notion of visual disparity to camera baselines featuring
six degrees of freedom (6 DoF) transformations, and present customized
cost volumes for this disparity.
\item We present a novel real-time and lightweight multi-level architecture
based on these cost volumes to perform end-to-end depth estimation
on video streams acquired in unstructured environments.
\item It is shown that M4Depth, is state of the art on the Mid-Air dataset~\cite{Fonder2019MidAir},
that it has good performances on the KITTI dataset~\cite{Geiger2012AreWe},
that it outperforms existing methods in a generalization setup on
the \tartanair dataset~\cite{Wang2020TartanAir}.
\end{enumerate}

\section{Problem statement\label{sec:problem-statement}}

First, we present the technicalities of the problem we want to solve.
We consider a camera rigidly attached to a vehicle moving within an
unknown static environment. The intrinsic parameters of the camera
are supposed to be known and constant. We introduce the following
components and notations:
\begin{itemize}
\item $\imSymbol_{\time}$ is an RGB image of size $\imHeight\times\imWidth$
recorded by the camera at time step $\time$. Images have the following
properties: 1) motion blur and rolling shutter artifacts are negligible;
2) the camera focal length $f$ is known and constant during a flight;
3) camera shutter speed and gain are unknown, and can change over
time.
\item $\transformationMatrix{\time}$ is the transformation matrix encoding
the motion of optical center of the camera from time step $\time-1$
to $\time$. This matrix is assumed to be known, which is realistic
when the camera motion is monitored such as with drones.
\item $\dpix_{\hImCoord\vImCoord,\time}$ is the $z$ coordinate (in meters)
of the point recorded by the pixel at coordinates $(\hImCoord,\vImCoord)$
of the frame $\imSymbol_{\time}$ with respect to the camera coordinate
system.
\end{itemize}
Using these notations, a depth map $\dmap_{\time}$ is an array of
$\dpix_{\hImCoord\vImCoord,\time}$ values with $\hImCoord j\in\{1,\ldots,\imWidth\}\times\{1,\ldots,\imHeight\}$.

We denote by $h_{\time}$ the complete series of image frames and
camera motions up to time step $\time$. We define a set $\mathcal{D}$
of functions $D$ that are able to estimate a depth map $\dmap$ from
$h_{\time}$, that is $\hat{\dmap_{\time}}=D(h_{\time})$, such that
$\ D\in\mathcal{D},\ \text{with}\ h_{\time}=\left[\imSymbol_{0},[\imSymbol_{1},\transformationMatrix 1],\ ...\ ,\ [\imSymbol_{\time},\transformationMatrix{\time}]\right]$.
Our objective is to find function $D$ in this set that best estimates
$\dmap_{\time}$.

Since collision avoidance is essential for autonomous vehicle applications,
errors in the estimate for closer objects should have a higher impact
than errors occurring for objects in the background of the scene.
This is taken care of by constructing a dedicated loss function for
training and by minimizing the error relatively to the distance of
the object. During testing, we will use the set of performance metrics
defined by Eigen~\etal to better grasp the behavior of our method.%

\section{Related work}

\label{sec:related-work}

Related works are presented according to four different categories.

\textbf{Depth from a single image.} Estimating depth from a single
RGB image is an old and well-established principle. If the first methods
were fully handcrafted~\cite{Saxena2006Learning}, the growth of
machine learning and the development of CNNs has impacted on the field
of depth estimation through the introduction of new methods such as
Monodepth~~\cite{Godard2017UnsupervisedMD}, Monodepth2~\cite{Godard2019DiggingIS},
or the method proposed by Pogi~\etal \cite{Pogi2018Towards} that
have led to massive improvements in the quality of the depth estimates.
Methods based on vision transformer networks, such as DPT~\cite{Ranftl2021Vision},
push the performance even further and are currently the state of the
art in the field.

Recent surveys~\cite{Ming2021DeepLearning,Xiaogang2020MonocularDE,Zhao2020MonocularDepth}
present summarized descriptions and comparisons of single image depth
estimation methods. The main observation made in these surveys is
that estimating depth from a single picture remains difficult, especially
for autonomous vehicle applications. Since the problem is ill-posed,
networks have to heavily rely on priors to compute a suitable proposal.
Such dependency on priors leads to a lack of robustness and generalization.
Therefore, methods of this family need to be fine-tuned for every
new scenario or environment encountered in order to produce good estimates.
Despite their massive parameter count, transformers are no exception
to these observations, as illustrated in \Figx{\ref{fig:graphical-abstract}}.

\textbf{Depth from an image sequence. }Methods exist that include
recurrence in networks to make use of temporal information for improving
estimates~\cite{Kumar2018DepthNetAR,Patil2020DontFT,Wang2019RecurrentNN,Wu2019Spatial,Zhang2019ExploitingTC}.
They are mainly adaptations of existing architectures, by adding or
modifying specific layers. As such, these methods do not make direct
use of motion information.

This issue has been thoroughly addressed by Watson~\etal~\cite{Watson2021TheTemporal}.
Their method, named ManyDepth, includes a model for motion and uses,
among other things, a cost volume built with the plane-sweeping method~\cite{Collins1996ASA,Gallup2007RealTimePS}.
In this method, both depth and motion are learned in an unsupervised
fashion. By ignoring the real camera motion, methods that work only
with sequences are unable to estimate the proper scale for depth without
relying on any prior knowledge about the structure of the scene. Even
worse, nothing prevents the scale estimated for the outputs to drift
over the sequence. This is problematic for autonomous vehicles moving
in unstructured environments.

\textbf{Use of motion information.} When motion is used by depth estimation
methods, it is mostly exploited to build a loss function for self-supervised
training~\cite{Godard2019DiggingIS,Luo2020EveryPC,Mahjourian2018Unsupervised,Patil2020DontFT,Wang2019RecurrentNN}.
In these cases, motion and depth are learnt by two independent networks
in an unsupervised fashion and depth is still estimated without any
clue about motion. As the core of these depth estimation networks
does not change compared to methods simply working on sequences, their
estimations suffer from the same issues as the ones produced by methods
that do not use motion.

One notable exception is the idea proposed by Luo~\etal~\cite{Luo2020ConsistentVD}.
Their method uses an self-supervised loss based on motion estimation
to fine-tune the network at test time. It achieves outstanding performance,
but at the cost of a large computational burden. Furthermore, this
method cannot estimate depth before the whole image sequence is available,
meaning that it only operates in an offline mode and makes it inappropriate
for autonomous vehicle applications.

\textbf{3D reconstruction.} Structure from motion (SfM) and multi-view
stereo (MVS) are two research fields that have developed in parallel
with depth estimation. The idea is to reconstruct 3D shapes from a
set of RGB images that capture the scene from different points of
view under specific hypothesis (MVS requires known camera poses, SfM
does not). Reconstruction is achieved by explicitly expressing the
relative camera position between the images of the set. Approaches
for performing this task are varied~\cite{Ozyesil2017ASurvey} and
are, by their nature, often unsuitable for real-time depth estimation.
However, some are adaptable for depth estimation on sequences~\cite{Gu2020Cascade,Ummenhofer2017DeMoNDA,Yao2018mvsnet},
while others are specifically designed to work on image sequences
in real time~\cite{Duzcceker2020DeepVideoMVSMS,Teed2020DeepV2DVT}.

The approach proposed by Duzcceker \etal~\cite{Duzcceker2020DeepVideoMVSMS}
and the method called DeepV2D by Teed \etal~\cite{Teed2020DeepV2DVT}
are similar. They both propose a three-stage network. Their stages
are an image-encoding network followed by the computation of a cost
volume that is finally processed by a depth estimation network. The
purpose of the cost volume consists of providing the costs for matching
a point in an image with a series of candidates in another image.
The cost volume of both methods is built by a plane-sweeping method~\cite{Collins1996ASA,Gallup2007RealTimePS}.

\begin{table}[t]
\caption{Main characteristics of a selection of depth estimation methods used
for comparison in this paper. \label{tab:methods_comp}}
\begin{adjustbox}{width=0.99\linewidth}
\centering{}%
\begin{tabular}{|l|c|c|c|c|c|}
\hline 
Method & Supervision & Multi-frame & Recurrent & Cam. pose & Pre-trained on \kitti\tabularnewline
\hline 
Monodepth~\cite{Godard2017UnsupervisedMD} & Self-sup. & No & No & No & Available\tabularnewline
Monodepth2~\cite{Godard2019DiggingIS} & Self-sup. & No & No & No & Available\tabularnewline
ST-CLSTM~\cite{Zhang2019ExploitingTC} & Self-sup. & No & Yes & No & Not available\tabularnewline
Wang~\cite{Wang2019RecurrentNN} & Self-sup. & No & Yes & No & Not available\tabularnewline
ManyDepth~\cite{Watson2021TheTemporal} & Self-sup. & Yes & No & Self-est. & Available\tabularnewline
DeepV2D~\cite{Teed2020DeepV2DVT} & Supervised & Yes & No & Self-est. & Available\tabularnewline
\hline 
M4Depth (ours) & Supervised & Yes & Yes & Given & N/A\tabularnewline
\hline 
\end{tabular}\end{adjustbox}
\end{table}

\textbf{Our baseline.} Based on this related work, we have selected
a representative set of existing methods for which the training code
is available, as given in \tabx{\ref{tab:methods_comp}}; they constitute
the baseline for our test bench. In this table, we indicate for each
method, respectively, the nature of its supervision mode, if it is
based on a single or multiple frames, if it is recurrent, how it deals
with the camera pose, and if weights for the \kitti are provided
by the authors.

\section{Description of M4Depth}

\label{sec:M4Depth-method}

Inspired by the previous works, we decided to base M4Depth on a multi-level
architecture trainable in an end-to-end fashion that relies on cost
volumes. The key novelty of M4Depth is the construction of a visual
disparity map for 6 DoF motion, that is converted into a depth map
by using motion information. This is described in the next section.

\subsection{Deriving depth from visual disparity}

\label{subsec:disp-equs}

In stereo vision, the relative pose between the cameras is fixed and
depth is estimated by calculating the disparity between the two camera
views. In multi-view approaches, the objective is either to calibrate
the cameras or to reconstruct a point cloud of the scene. In our setup,
the camera moves along a path known thanks to the on-board inertial
measurement unit. Since we only know the relative motion between two
camera locations which is not fixed, we have to define an appropriate
notion of visual disparity to be able to estimate a dense depth map
from the pixelwise frame-to-frame pixel displacement vectors. By doing
so, the network only has to estimate the visual disparity to compute
depth.

Our notion of visual disparity, new to the best of our knowledge,
denoted by $\rho$ is established as follows. The transformation matrix
$\transformationMatrix{\time}$ formalizing the known physical camera
motion with 6 DoF between consecutive frames of the video stream can
be decomposed as a rotation matrix $\mathbf{R}_{\time}$ and a 3D
translation vector $\translationVector_{\time}$. Using the classical
pinhole camera model, a point $P$ in space seen by the camera at
two different time instants $\time$ and $\time-1$, and projected
at coordinates $(\hImCoord_{\secondIndex},\vImCoord_{\secondIndex})$
in the current frame $t$ is linked to its previous coordinates $(\hImCoord_{\firstIndex},\vImCoord_{\firstIndex})$
in frame at time $\time-1$ by the motion $\transformationMatrix{\time}$
as follows
\begin{equation}
\dpix_{\hImCoord_{t-1}\vImCoord_{t-1}}\begin{bmatrix}\hImCoord_{\firstIndex}\\
\vImCoord_{\firstIndex}\\
1
\end{bmatrix}=\mathbf{K}\left(\rotMatrix{\time}\dpix_{\hImCoord_{\time}\vImCoord_{\time}}\begin{bmatrix}\hImCoord_{\secondIndex}/\focal x\\
\vImCoord_{\secondIndex}/\focal y\\
1
\end{bmatrix}+\translationVector_{\time}\right)\comma\label{equ:pinhole}
\end{equation}

where $\dpix_{\hImCoord_{\time}\vImCoord_{\time}}=\dmap_{\time}(\hImCoord_{t},\vImCoord_{t})$
is the depth of the point $P$ at time $\time$, and $\mathbf{K}$
is a camera calibration matrix, which is identical at times $\time$
and $\time-1$. It is reasonable to simplify the expression of the
$3\times3$ $\mathbf{K}$ matrix to 
\begin{equation}
\mathbf{K}=\text{diag}(\focal x,\focal y,1)\comma
\end{equation}
with $\focal x$ and $\focal y$ being the focal lengths along the
$x$ and $y$ axes respectively. This assumes that the coordinates
$(i,j)$ are expressed directly with respect to the principal point
of the sensor $(c_{x},c_{y})$ and that the skew parameter is negligible.

Before defining our visual disparity, we rewrite \Eqx{\ref{equ:pinhole}}
as 
\begin{equation}
\dpix_{\hImCoord_{t-1}\vImCoord_{t-1}}\begin{bmatrix}\hImCoord_{\firstIndex}\\
\vImCoord_{\firstIndex}\\
1
\end{bmatrix}=\dpix_{\indexOfVirtualCamera}\ \dpix_{\hImCoord_{\time}\vImCoord_{\time}}\begin{bmatrix}\hImCoord_{\indexOfVirtualCamera}\\
\vImCoord_{\indexOfVirtualCamera}\\
1
\end{bmatrix}+\begin{bmatrix}\focal x\translationVectorComp_{x}\\
\focal y\translationVectorComp_{y}\\
\translationVectorComp_{z}
\end{bmatrix}\comma\label{equ:substitute}
\end{equation}
with 
\begin{equation}
[\dpix_{\indexOfVirtualCamera}\hImCoord_{\indexOfVirtualCamera},\ \dpix_{\indexOfVirtualCamera}\vImCoord_{\indexOfVirtualCamera},\ \dpix_{\indexOfVirtualCamera}]^{\transpositionSymbol}=\ \mathbf{K}\ \mathbf{R}\ [\hImCoord_{\secondIndex}/\focal x,\ \vImCoord_{\secondIndex}/\focal y,\ 1]^{\transpositionSymbol}\point
\end{equation}
From this equation, we can see that $(\hImCoord_{\indexOfVirtualCamera},\vImCoord_{\indexOfVirtualCamera})$
are the coordinates of the point $P$ in the plane of a virtual camera
$\indexOfVirtualCamera$ whose origin is the same as the camera at
time $\time$ but with the orientation of the camera at time $\time-1$.

We now define our pixelwise visual disparity $\disp_{\hImCoord_{\time}\vImCoord_{\time}}$
as the Euclidean norm 
\begin{equation}
\disp_{\hImCoord_{\time}\vImCoord_{\time}}=\sqrt{\Delta_{\hImCoord_{\time}}^{2}+\Delta_{\vImCoord_{\time}}^{2}}\label{eq:visual_disparity_}
\end{equation}
where 
\[
\left[\begin{array}{c}
\Delta_{\hImCoord_{\time}}\\
\Delta_{\vImCoord_{\time}}
\end{array}\right]=\left[\begin{array}{c}
\hImCoord_{\firstIndex}-\hImCoord_{\indexOfVirtualCamera}\\
\vImCoord_{\firstIndex}-\vImCoord_{\indexOfVirtualCamera}
\end{array}\right]\,.
\]

After reorganization, using \Eqx{\ref{equ:substitute}} and simplification,
we get 
\begin{equation}
\begin{bmatrix}\Delta_{\hImCoord_{\time}}\\
\Delta_{\vImCoord_{\time}}
\end{bmatrix}=\frac{1}{\dpix_{\hImCoord_{\time}\vImCoord_{\time}}\ \dpix_{\indexOfVirtualCamera}+\translationVectorComp_{z}}\begin{bmatrix}\focal x\translationVectorComp_{x}-\translationVectorComp_{z}\hImCoord_{\indexOfVirtualCamera}\\
\focal y\translationVectorComp_{y}-\translationVectorComp_{z}\vImCoord_{\indexOfVirtualCamera}
\end{bmatrix}\point\label{equ:flow}
\end{equation}

By taking into account the physics of a scene and the camera motion
of an autonomous vehicle, it can be shown that $\dpix_{\hImCoord_{\time}\vImCoord_{\time}}\ \dpix_{\indexOfVirtualCamera}+\translationVectorComp_{z}$
should rarely be negative. As a result, the disparity $\disp_{\hImCoord_{\time}\vImCoord_{\time}}$
can be computed as follows 
\begin{equation}
\begin{aligned}\disp_{\hImCoord_{\time}\vImCoord_{\time}}= & \frac{\sqrt{\left(\focal x\translationVectorComp_{x}-\translationVectorComp_{z}\hImCoord_{\indexOfVirtualCamera}\right)^{2}+\left(\focal y\translationVectorComp_{y}-\translationVectorComp_{z}\vImCoord_{\indexOfVirtualCamera}\right)^{2}}}{\dpix_{\hImCoord_{\time}\vImCoord_{\time}}\ \dpix_{\indexOfVirtualCamera}+\translationVectorComp_{z}}\point\end{aligned}
\label{equ:disp2depth}
\end{equation}

This expression links the disparity for a pixel to the depth of the
corresponding point in space. Since disparity can be estimated from
the RGB content of two consecutive images, we have a mean to estimate
the depth by inverting the equation, yielding 
\begin{equation}
\dpix_{\hImCoord_{\time}\vImCoord_{\time}}=\frac{\sqrt{\left(\focal x\translationVectorComp_{x}-\translationVectorComp_{z}\hImCoord_{\indexOfVirtualCamera}\right)^{2}+\left(\focal y\translationVectorComp_{y}-\translationVectorComp_{z}\vImCoord_{\indexOfVirtualCamera}\right)^{2}}}{\disp_{\hImCoord_{\time}\vImCoord_{\time}}\ \dpix_{\indexOfVirtualCamera}}-\frac{\translationVectorComp_{z}}{\dpix_{\indexOfVirtualCamera}}\point\label{equ:disp_to_depth}
\end{equation}

In practice, there are different ways to estimate $\disp_{\hImCoord_{\time}\vImCoord_{\time}}$,
and in our method, we build various proposals for $\disp_{\hImCoord_{\time}\vImCoord_{\time}}$
and let the network use them to compute the best estimate. Note that,
once a disparity map $\dispmap_{t}$ has been estimated, the $(\hImCoord_{\firstIndex},\vImCoord_{\firstIndex})$
coordinates are given by a function $\Psi$, parametrized as follows

\begin{equation}
(\hImCoord_{\firstIndex},\vImCoord_{\firstIndex})=\Psi(\hImCoord_{\secondIndex},\vImCoord_{\secondIndex},\transformationMatrix{\time},\dispmap_{t})\point\label{eq:mapping-with-ijTr}
\end{equation}
These $(\hImCoord_{\firstIndex},\vImCoord_{\firstIndex})$ coordinates
are defined on a continuous space instead of a discrete grid.

\subsection{Definition of the network}

To find the disparity, we could resort to the point triangulation
process, which is by nature an iterative process. Instead of simply
iterating on a full network as proposed in~\cite{Ummenhofer2017DeMoNDA}
or bypassing the iteration by proposing a full range of candidates
as in~\cite{Duzcceker2020DeepVideoMVSMS,Teed2020DeepV2DVT}, we approach
the iterative process as a multi-scale pyramidal network, as PWC-Net~\cite{Sun2018PWCNetCF}.
By doing so, the iterative process is embedded in the architecture
itself. This is an adaptation of the U-Net encoder-decoder architecture
with skip connections~\cite{Ronneberger2015UNet}, where each level
$\levelIndex$ of the decoder has to produce an estimate for the desired
output, that is a disparity map in our case. In the decoder, the estimate
produced at one level is forwarded to feed the next level to be refined.
The levels of this type of architecture are generic and can be stacked
at will to obtain the desired network depth.

\begin{figure*}[!t]
\begin{centering}
\includegraphics[width=0.9\linewidth]{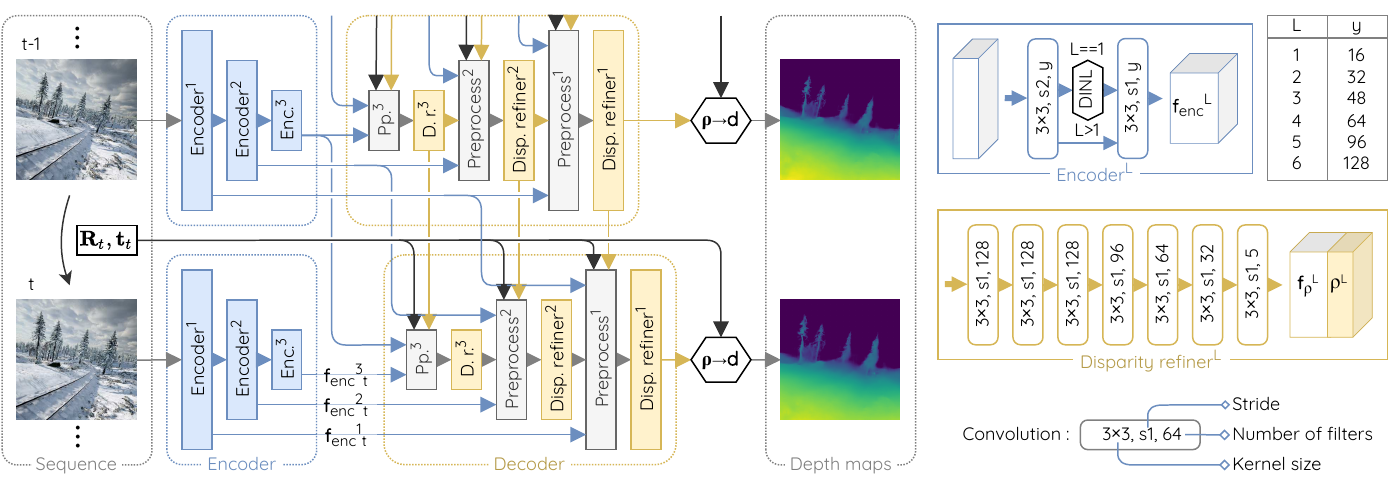}
\par\end{centering}
\centering{}\caption{Architecture overview of M4Depth (with three levels here), fed by
two consecutive frames and the camera motion. Each disparity refiner
produces a disparity estimate and learnable disparity features. All
convolutions are followed by a leaky ReLU activation unit~\cite{Xu2015EmpiricalEO},
except for the ones producing a disparity estimate. To ease the convergence,
disparities are encoded in the log-space.\label{fig:global-overview}}
\end{figure*}

Our architecture, illustrated in \Figx{\ref{fig:global-overview}},
uses the same standard encoder network as PWC-Net~\cite{Sun2018PWCNetCF}
with the only exception that we add a Domain-Invariant Normalization
layer (DINL)~\cite{Zhang2020Domain} after the first convolutional
layer. We use it to increase the robustness of the network to varied
colors, contrasts and luminosity conditions without increasing the
number of convolutional filters. 

At each level $L$ of the decoder, a small convolutional subnetwork
is in charge of refining the disparity map. We named it the disparity
refiner. Its inputs are the upscaled disparity estimate made by the
previous decoder level in the architecture and a series of preprocessed
data generated by a preprocessing unit.

\begin{figure}[t!]
\centering{}\includegraphics[width=0.9\linewidth]{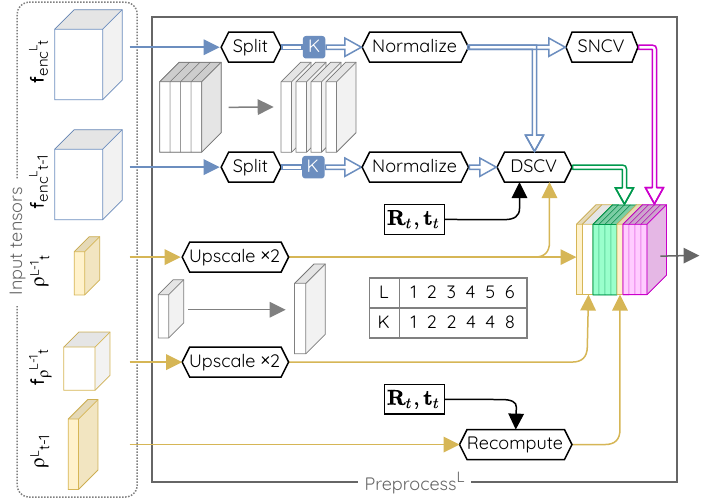}\caption{Details of the operations performed by our preprocessing units. They
do not feature any learnable parameters. The split layer subdivides
feature vectors in K sub-vectors to be processed in parallel in subsequent
steps. We give the value of K that we use for an architecture with
six levels.\label{fig:level-overview}}
\end{figure}

The preprocessing unit is illustrated in \Figx{\ref{fig:level-overview}}.
It is made of fixed operations and has no learnable parameters. Its
purpose is to prepare the input for the next disparity refiner. 
In short, the preprocessor has two main purposes. First, it adapts
the vectors of the feature maps produced by the encoders  to make
the network robust to unknown visual inputs. For that, it uses this
data alongside camera motion to build two distinct cost volumes, the
Disparity Sweeping Cost Volume (DSCV) and the Spatial Neighborhood
Cost Volume (SNCV). Second, it recomputes the disparity estimate obtained
for the previous time by adjusting it to the camera motion. These
data are then concatenated and forwarded to the disparity refiner.

Let us now detail these operations and motivate their use. %

\textbf{Split and normalize layers.} The use of leaky ReLU activation
units in the encoder can lead to a feature maps containing plenty
of small values. While classification or segmentation networks rely
on the raw value of each entry in a feature vector, our network relies
on the relative differences between neighboring feature vectors through
the use of cost volumes. To get good generalization properties, this
relative difference should remain significant in all situations. The
split and normalize layers ensure that this is the case.

The split layer subdivides feature vectors in K sub-vectors to be
processed in parallel in subsequent layers. It gives the network the
ability to decouple the relative importance of specific features within
a same vector by assigning them to different sub-vectors.

The normalize layer normalizes the features of a same sub-vector and
therefore levels the difference in magnitude of different sub-vectors.
This is beneficial for the disparity refiner layers as this normalization
leads the outputs of the cost volumes to span to a known pre-defined
range. It also allows a full use of the information embedded in sub-vectors
whose magnitude is very small because of the leaky ReLU activation
units.

\textbf{Recompute layer.} The disparities estimated by the network
are specific to the motion occurring between two given frames. By
using the set of equations developed in Section~\ref{subsec:disp-equs}
and if the camera motion is known, it is possible to compute the disparities
that should be observed at a given time step from a previous disparity
estimate. The purpose of the recompute layer is to update the disparities
estimated for the previous frame to provide an hint in the form of
a first estimate of the disparities for the current frame.

\textbf{Spatial Neighborhood Cost Volume (SNCV).} This cost volume
is computed from a single feature map $\mathbf{f}$ and is a form
of spatial autocorrelation. Each pixel of the cost volume is assigned
the costs of matching the feature vector located at the same location
in the feature map with the neighboring feature vectors located within
a given range $r$ of the considered location 
\begin{equation}
\begin{aligned} & \text{SNCV}_{r}(\mathbf{f})(\hImCoord,\vImCoord)=\\
 & \left[\ \text{cost}\left(\mathbf{f}(\hImCoord,\vImCoord),\mathbf{f}(\hImCoord+p,\vImCoord+q)\right)\;\forall\;p,q\in\{-r,\ldots,r\}\ \right]\comma
\end{aligned}
\end{equation}
where the cost of matching two vectors $\textbf{x}_{1}$ and $\textbf{x}_{2}$
of dimension $N$ is defined as their correlation~\cite{Dosovitskiy2015FlowNet,Xu2017Accurate}
\begin{equation}
\text{cost}(\textbf{x}_{1},\textbf{x}_{2})=\frac{1}{N}\textbf{x}_{1}^{\transpositionSymbol}\textbf{x}_{2}\point
\end{equation}
The SNCV gives an indication about the two-dimensional spatial structure
of the scene captured by the features of the encoder. By design, it
is impossible to recover the feature vectors that led to a given cost
value. Network parameters trained with this cost metric will therefore
be invariant to changes in the input feature vectors if they lead
to the same cost value. This can help to obtain a robust and generalizable
depth estimation network, which was not achievable by forwarding the
feature map directly.

\textbf{Disparity Sweeping Cost Volume (DSCV).} This cost volume is
computed from two consecutive feature maps $\mathbf{f}_{\time-1}$
and $\mathbf{f}_{\time}$, and a disparity map estimate $\dispmapEstimate_{\time}$
(see left of \Figx{\ref{fig:level-overview}}). For each pixel,
the cost volume assigns the cost of matching the feature vector located
at the same place in $\mathbf{f}_{\time}$ with the corresponding
reprojected feature vectors from $\mathbf{f}_{\time-1}$ according
to
\begin{equation}
\begin{aligned} & \text{DSCV}_{\delta}(\mathbf{f}_{\time},\mathbf{f}_{\time-1},\dispmapEstimate_{\secondIndex})(\hImCoord,\vImCoord)=\\
 & \left[\text{cost}\left(\mathbf{f}_{\time}(\hImCoord,\vImCoord),\mathbf{f}_{\text{reproj}}(\hImCoord,\vImCoord,\Delta_{\rho})\right)\;\forall\;\Delta_{\rho}\in\{-\delta,\ldots,\delta\}\right]\point
\end{aligned}
\end{equation}
In that expression, the feature map $\mathbf{f}_{\time-1}$ is reprojected
for a range $\delta$ of disparities equally distributed around a
given estimate, that is 
\begin{equation}
\mathbf{f}_{\text{reproj}}(\hImCoord,\vImCoord,\Delta_{\rho})=\mathbf{f}_{\time-1}\left(\Psi\left(\hImCoord,\vImCoord,\transformationMatrix{\time},\max\left(\epsilon,\dispmapEstimate_{\secondIndex}+\Delta_{\rho}\right)\right)\right)\comma
\end{equation}
where $\Psi$ is given by \Eqx{\ref{eq:mapping-with-ijTr}}, and
$\epsilon>0$. In that expression, $\mathbf{f}_{\text{reproj}}$ is
interpolated from $\mathbf{f}_{\time-1}$ since $\Psi$ returns real
coordinates , and $\max\left(\epsilon,\dispmapEstimate_{\secondIndex}+\Delta_{\rho}\right)$
ensures the positiveness of the disparity used for computing the reprojection.
Each vector element of the cost volume corresponds to one given disparity
correction with respect to the provided estimate. Browsing through
a range of disparities for each pixel creates a series of candidates
for the corresponding reprojected point. By searching for the reprojected
candidate that is the most similar to the visual information observed
at time step $t$, it is possible to assess which disparity is the
most likely to be associated to each pixel.

Since we are using a pyramidal architecture, a range $\delta$ at
the $L$-th level is equivalent to a range $\delta2^{L}$ in the original
picture. As opposed to methods directly using depth for building a
cost volume, this property helps with scanning a larger range of disparities
while keeping a pixelwise precision in the final picture without having
to check a large number of candidates.

\textbf{}%

\subsection{Loss function definition}

Since the levels of our architecture are stackable at will, the architecture
can have any depth. We now detail our loss function for a network
that is made of $M$ levels.

As in previous works~\cite{Chen2019Structure,Ligin2019Pyramid,Liu2021MultiScale},
we use a multi-scale loss function. For each frame and each level,
we compute the $L_{1}$ distance on the logarithm of the depths resulting
from the conversion of disparity estimates using \Eqx{\ref{equ:disp_to_depth}}.
The logarithm leads to a scale invariant loss function~\cite{Eigen2014Depth}
and the use of an $L_{1}$ distance is motivated by its good convergence
properties~\cite{Carvalho2018OnRegression}. Since intermediate depth
maps have a lower resolution, ground truths are resized by bilinear
interpolation to match the dimensions of the estimates. The resulting
terms are aggregated through a weighted sum, yielding 
\begin{equation}
\mathcal{L}_{t}=\frac{1}{\imHeight\imWidth}\sum_{\levelIndex=1}^{M}\sum_{\dpix_{ij}\in\dmap_{t}^{l}}2^{\levelIndex+1}\ \left|\log(\dpix_{ij})-\log(\hat{\dpix}_{ij})\right|\point
\end{equation}

The total loss for the sequence is defined as the average of the loss
computed for each of its time steps.

\section{Experiments\label{sec:Experiments}}

In this section, we present three experiments to analyze the performance
of our M4Depth method. For each of them, we detail the chosen dataset,
the training (if appropriate), and discuss the results. Performance
metrics are taken from Eigen \etal~\cite{Eigen2014Depth} for depth
maps limited to 80~$m$.

In order to compare our method with its parent, PWC-Net~\cite{Sun2018PWCNetCF},
we perform the same experiments for both M4Depth and PWC-Net. As PWC-Net
is an optical flow network, we use \Eqx{\ref{equ:pinhole}} to get
the frame-to-frame optical flow from depth and motion for training
the network. During testing, we estimate depth from the optical flow
by first computing our visual disparity from the estimated optical
flow.

\subsection{Unstructured environments}

For our first experiment, we compare the performance of our method
with the ones of the state of the art on a dataset featuring unstructured
environments.

\textbf{\midair} \textbf{dataset.} For this experiment, we use Mid-Air~\cite{Fonder2019MidAir}.
This synthetic dataset consists of a set of drone trajectories recorded
in large, unstructured and static environments under varied weather
and lighting conditions. All trajectories are recorded in different
places of the environments, which means that there is little overlap
between the visual content of two individual trajectories. This allows
one to build a test set whose content is not present in the training
set while belonging to the same data distribution. In addition, \midair
meets all the assumptions of our problem statement (see \secx{\ref{sec:problem-statement}}),
which makes it a perfect choice.

The first performance reported on \midair for depth estimation was
provided recently by Miclea~\etal~\cite{Miclea2022Monocular}.
The authors of this paper do, however, not provide the details required
to reproduce their train and test splits and their results. As a result,
we have to define our own splits.

The dataset features $192$ trajectories. We select one in three to
create our test set, which is more varied than the small test set
suggested in the original paper~\cite{Fonder2019MidAir}. The frame
rate is subsampled by a factor four (from $25$ to $6.25$~fps) to
increase the apparent motion between two frames. For all our experiments,
images and depth maps are resized to a size of $384\times384$ pixels.
We use bilinear interpolation to resize color images and the nearest
neighbor method for depth maps.

\textbf{Training. }We use the He initialization~\cite{He2015DelvingDI}
for our variables and the Adam optimizer~\cite{Kingma2015Adam} for
the training itself. We keep the default parameters proposed by both
of them. The learning rate is set to $10^{-4}$. We trained our network
with six levels. All our trainings are performed on sequences of four
time steps and with a batch size of three sequences. The network is
trained on a GPU with $16~\text{GB}$ of VRAM for $220~\text{k}$
iterations. After each epoch we compute the performance of the network
on the validation set of the \kitti dataset to avoid any overfitting,
and keep a copy of the best set of weights to be used for the tests
after the training.

A series of data augmentation steps are performed on each sequence
during the training to boost the robustness of our trained network
to visual novelties. More precisely, we apply the same random brightness,
contrast, hue, and saturation change to all the RGB pictures of a
sequence and the colors of a sequence are inverted with a $50\%$
probability. Finally, we randomly rotate the data of the sequence
by a multiple of $90$~degrees around the $z$-axis of the camera
when training on Mid-Air. With these settings, a training requires
approximately $30$ hours.

Because of the lack of reproducible performances reported on \midair,
we had to train a selection of state-of-the-art methods drawn in \tabx{\ref{tab:methods_comp}}
to build a baseline. The training details for the chosen methods are
given in the supplementary material. We could not guarantee to get
the best performance out of DeepV2D~\cite{Teed2020DeepV2DVT} because
of the importance of its hyper-parameters and the excessive duration
of its training time. We, therefore, decided to discard it for this
experiment.

\textbf{Results.} The results are reported in \tabx{\ref{tab:perf_scores-midair}}.
In this table and following tables, the best score for a metric is
highlighted in bold and the second best is underlined.
\begin{table*}[!p]
\centering{}\caption{Performance comparison on our test set of \midair. Here, a 6-level
version of M4Depth is compared to the baseline methods. Scores correspond
to the best performance obtained out of five individual network trainings.
The best score for a metric is highlighted in bold and the second
best is underlined. \label{tab:perf_scores-midair}}
\begin{tabular}{|l|c||c|c|c|c||c|c|c|}
\hline 
Method & Test size & \cellcolor{blue!25}Abs Rel $\downarrow$ & \cellcolor{blue!25}SQ Rel $\downarrow$ & \cellcolor{blue!25}RMSE $\downarrow$ & \cellcolor{blue!25}RMSE log $\downarrow$ & \cellcolor{orange!40}$\delta<1.25$ $\uparrow$ & \cellcolor{orange!40}$\delta<1.25^{2}$ $\uparrow$ & \cellcolor{orange!40}$\delta<1.25^{3}$ $\uparrow$\tabularnewline
\hline 
Monodepth~\cite{Godard2017UnsupervisedMD} & $384\times384$ & 0.314 & 8.713 & 13.595 & 0.438 & 0.678 & 0.828 & 0.895\tabularnewline
Monodepth2~\cite{Godard2019DiggingIS} & $384\times384$ & 0.394 & 5.366 & 12.351 & 0.462 & 0.610 & 0.751 & 0.833\tabularnewline
ST-CLSTM~\cite{Zhang2019ExploitingTC} & $384\times384$ & 0.404 & 6.390 & 13.685 & 0.438 & \uline{0.751} & 0.865 & 0.911\tabularnewline
Wang~\cite{Wang2019RecurrentNN} & $384\times384$ & 0.241 & 5.532 & 12.599 & 0.362 & 0.648 & 0.831 & 0.911\tabularnewline
ManyDepth~\cite{Watson2021TheTemporal} & $384\times384$ & 0.203 & 3.549 & 10.919 & 0.327 & 0.723 & \uline{0.876} & \uline{0.933}\tabularnewline
\hline 
PWCDC-Net~\cite{Sun2018PWCNetCF} & $384\times384$ & \textbf{0.095} & \textbf{2.087} & \uline{8.351} & \uline{0.215} & 0.887 & 0.938 & 0.962\tabularnewline
M4Depth-d6 (Ours) & $384\times384$ & \uline{0.105} & \uline{3.454} & \textbf{7.043} & \textbf{0.186} & \textbf{0.919} & \textbf{0.953} & \textbf{0.969}\tabularnewline
\hline 
\end{tabular}
\end{table*}

\begin{figure*}[t]
\begin{adjustbox}{width=0.9\linewidth}%
\begin{minipage}[t]{1.1\linewidth}%
\setlength{\tabcolsep}{2pt}%
\begin{tabular}{cccccccc}
\includegraphics[width=0.13\textwidth]{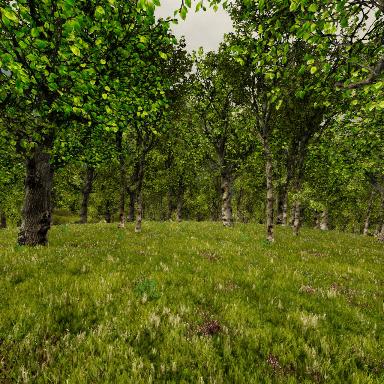} & \includegraphics[width=0.13\textwidth]{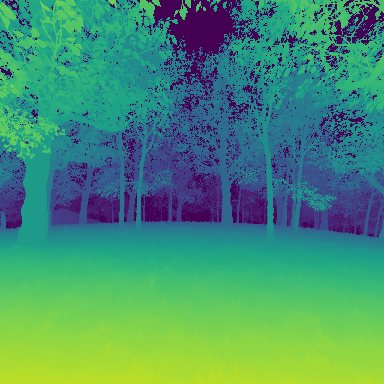} & \includegraphics[width=0.13\textwidth]{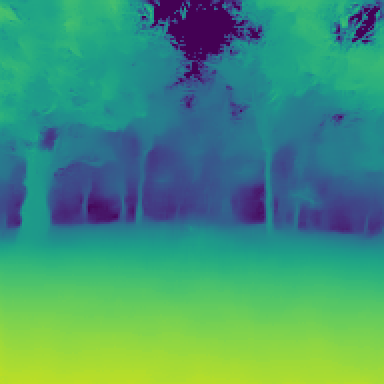} & \includegraphics[width=0.13\textwidth]{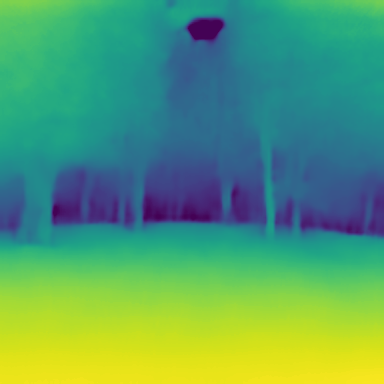} & \includegraphics[width=0.13\textwidth]{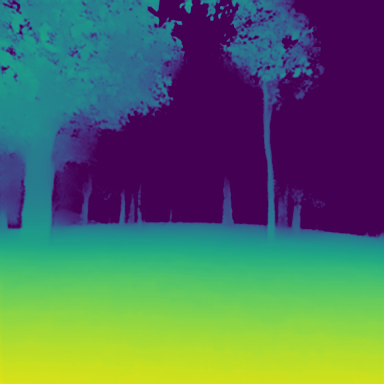} & \includegraphics[width=0.13\textwidth]{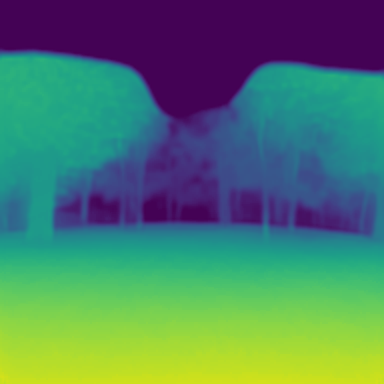} & \includegraphics[width=0.13\textwidth]{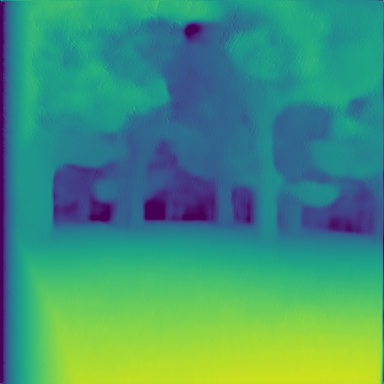} & \includegraphics[width=0.13\textwidth]{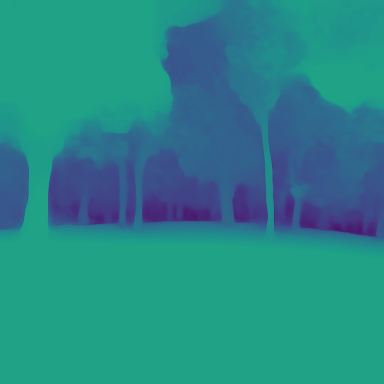}\tabularnewline
\includegraphics[width=0.13\textwidth]{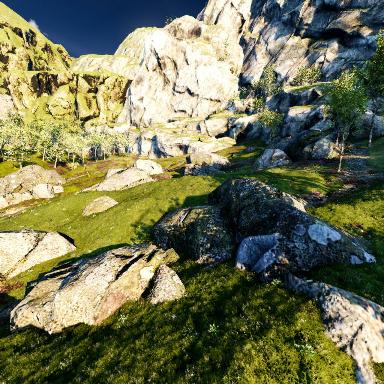} & \includegraphics[width=0.13\textwidth]{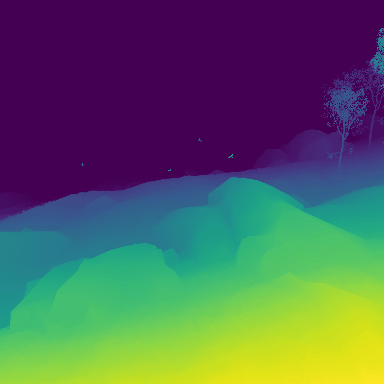} & \includegraphics[width=0.13\textwidth]{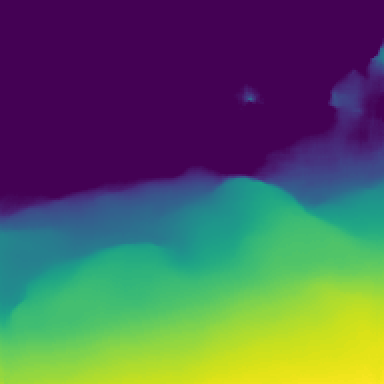} & \includegraphics[width=0.13\textwidth]{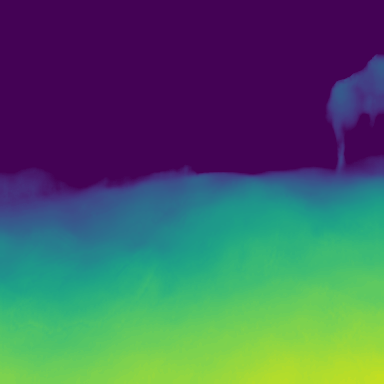} & \includegraphics[width=0.13\textwidth]{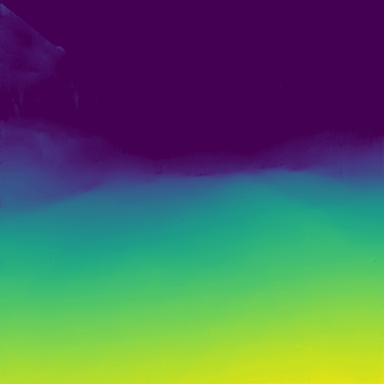} & \includegraphics[width=0.13\textwidth]{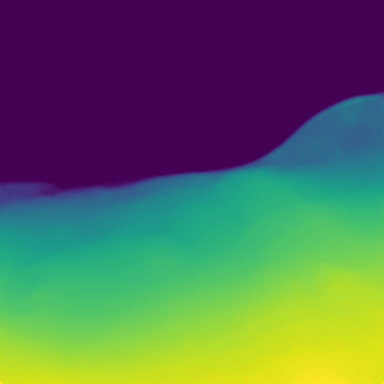} & \includegraphics[width=0.13\textwidth]{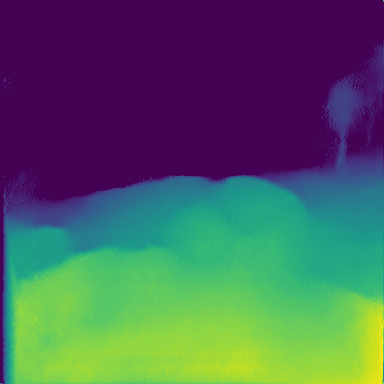} & \includegraphics[width=0.13\textwidth]{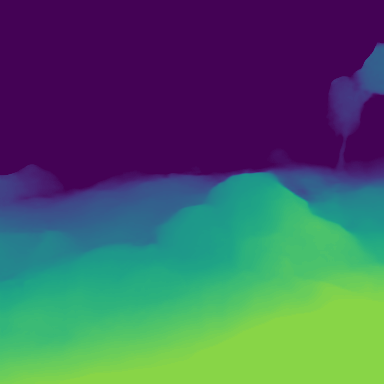}\tabularnewline
\includegraphics[width=0.13\textwidth]{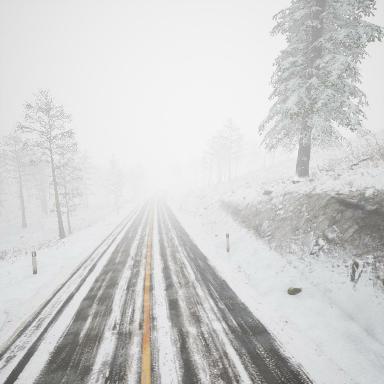} & \includegraphics[width=0.13\textwidth]{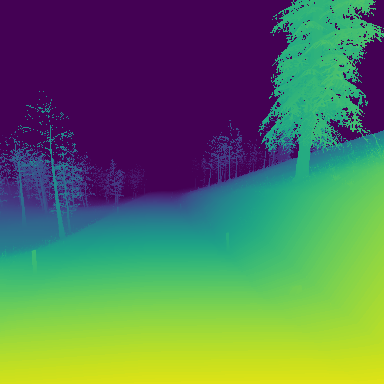} & \includegraphics[width=0.13\textwidth]{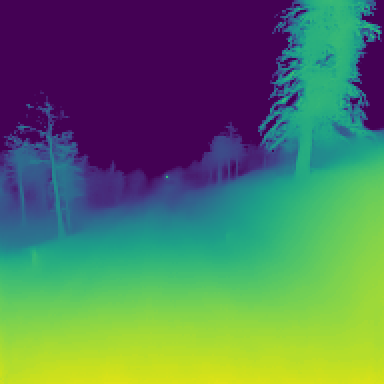} & \includegraphics[width=0.13\textwidth]{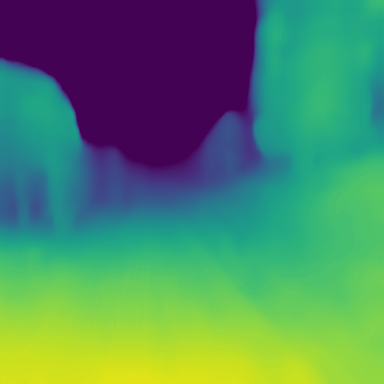} & \includegraphics[width=0.13\textwidth]{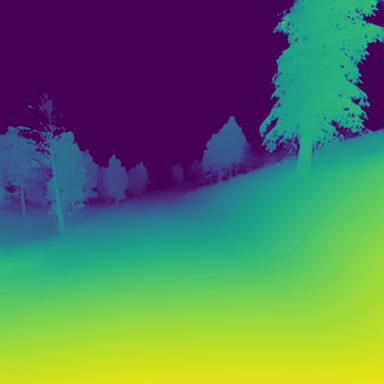} & \includegraphics[width=0.13\textwidth]{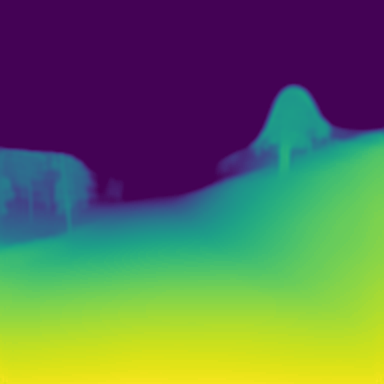} & \includegraphics[width=0.13\textwidth]{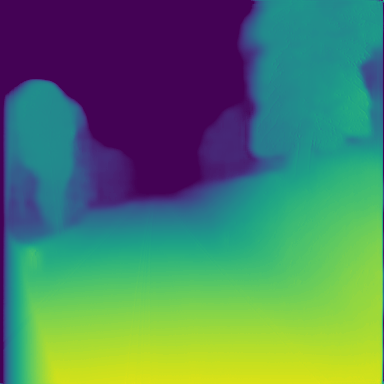} & \includegraphics[width=0.13\textwidth]{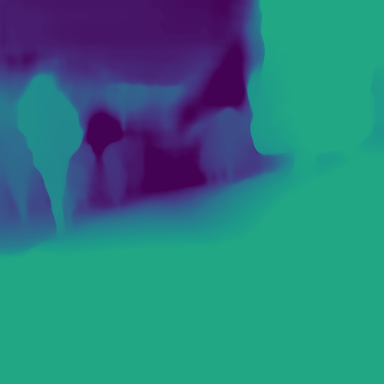}\tabularnewline
\vspace{-0.8cm}
 &  &  &  &  &  &  & \tabularnewline
\subfloat[\foreignlanguage{english}{RGB image}]{\hspace{0.121\textwidth}

} & \subfloat[\foreignlanguage{english}{Ground truth}]{\hspace{0.121\textwidth}

} & \subfloat[\foreignlanguage{english}{M4Depth (ours)}]{\hspace{0.121\textwidth}

} & \subfloat[\foreignlanguage{english}{ManyDepth~\cite{Watson2021TheTemporal}}]{\hspace{0.121\textwidth}

} & \subfloat[\foreignlanguage{english}{Wang~\cite{Wang2019RecurrentNN}}]{\hspace{0.121\textwidth}

} & \subfloat[\foreignlanguage{english}{ST-CLSTM\cite{Zhang2019ExploitingTC}}]{\hspace{0.121\textwidth}

} & \subfloat[\foreignlanguage{english}{Monodepth~\cite{Godard2017UnsupervisedMD}}]{\hspace{0.121\textwidth}

} & \subfloat[\foreignlanguage{english}{Monodepth2~\cite{Godard2019DiggingIS}}]{\hspace{0.121\textwidth}

}\tabularnewline
\end{tabular}%
\end{minipage}\end{adjustbox}\caption{Comparison of the depth maps estimated by M4Depth and baseline methods.
M4Depth recovers depth details more accurately than baseline methods.\label{fig:qualitative-overview}}
\end{figure*}

Globally, it appears that M4Depth outperforms the baseline methods.
However, it slightly underperforms on the relative performance metrics
when compared to PWC-Net. This observation, compared with the excellent
performances on other metrics, indicates that our network tends to
overestimate depth more often than other methods. A qualitative comparison
of the outputs of the different methods is shown in \Figx{\ref{fig:qualitative-overview}}.
From this figure, we observe that although M4Depth lacks of details
in areas with sharp depth transitions, it recovers depth details more
accurately than baseline methods, even for challenging scene elements
such as forests or unstructured terrain.

\subsection{Standard depth estimation benchmark}

The purpose of the second experiment is to assess the performance
on a standard depth estimation benchmark.

\textbf{\kitti dataset}~\cite{Geiger2012AreWe}\textbf{.} Most real
datasets that provide RGB+D and motion data focus on cars driving
in partially dynamic urban environments~\cite{Geiger2012AreWe,Ros2016TheSynthia,Sturm2012ABenchmark}.
In this field, \kitti is the reference benchmark dataset when evaluating
the performance of a depth estimation method. \kitti is not fully
compliant with our problem statement: it has incomplete depth maps,
there are some moving objects, and the camera has only three degrees
of freedom, etc. Despite that, it is a good practical choice for doing
tests on real data.

We use the dataset split proposed by Eigen~\etal~\cite{Eigen2014Depth}.
The camera pose is estimated by a combined GPS-inertial unit and is
therefore subject to measurement imperfections. Since a few samples
were recorded in urban canyons where poor GPS reception induced erratic
pose estimates, and as our method requires reliable pose estimates,
we discarded these problematic samples from the splits. Additionally,
we also subsampled the frame rate by a factor of two (from $10$ to
$5$~fps) to roughly match the one of our \midair sets. Finally,
images were resized to $256\times768$ pixels.

\textbf{Training. }For tests on \kitti, we reuse the weights of the
network with 6 levels trained on \midair and fine-tune them for $20\,\text{k}$
additional iterations on a $50-50\%$ mix of \kitti and \midair
samples. The fine-tuning is required to train our network to deal
with large areas with poor textures and frame-to-frame illumination
changes as these characteristics are not present in \midair. As the
ground-truth depth maps for \kitti were generated from Lidar measurements,
they are sparse and fine details are missing in the ground truths.
Shortcomings created by these imperfections can be mitigated by fine-tuning
on both datasets. During the fine-tuning, we also perform random color
augmentation on the sequences. With these settings, the fine-tuning
requires three hours.

\textbf{Results.} The performance of M4Depth with six levels on the
\kitti dataset is reported in \tabx{\ref{tab:perf_scores-kitti}}.
\begin{table*}[p]
\begin{centering}
\caption{Performance of M4Depth (best of 5 trainings) on the \kitti dataset
. The scores reported for reference methods are the ones published
by their respective authors. \label{tab:perf_scores-kitti}}
\par\end{centering}
\centering{}%
\begin{tabular}{|l|c||c|c|c|c||c|c|c|}
\hline 
Method & Test size & \cellcolor{blue!25}Abs Rel $\downarrow$ & \cellcolor{blue!25}SQ Rel $\downarrow$ & \cellcolor{blue!25}RMSE $\downarrow$ & \cellcolor{blue!25}RMSE log $\downarrow$ & \cellcolor{orange!40}$\delta<1.25$ $\uparrow$ & \cellcolor{orange!40}$\delta<1.25^{2}$ $\uparrow$ & \cellcolor{orange!40}$\delta<1.25^{3}$ $\uparrow$\tabularnewline
\hline 
Monodepth~\cite{Godard2017UnsupervisedMD} & $256\times512$ & 0.114 & 0.898 & 4.935 & 0.206 & 0.861 & 0.949 & 0.976\tabularnewline
Monodepth2~\cite{Godard2019DiggingIS} & $320\times1024$ & 0.106 & 0.806 & 4.630 & 0.193 & 0.876 & 0.958 & 0.980\tabularnewline
ST-CLSTM~\cite{Zhang2019ExploitingTC} & $375\times1240$ & 0.104 & N/A & 4.139 & 0.131 & 0.833 & 0.967 & 0.988\tabularnewline
ManyDepth~\cite{Watson2021TheTemporal} & $320\times1024$ & 0.087 & 0.685 & 4.142 & 0.167 & 0.920 & 0.968 & 0.983\tabularnewline
Wang~\cite{Wang2019RecurrentNN} & $128\times416$ & \uline{0.077} & \uline{0.205} & \textbf{1.698} & \uline{0.110} & \uline{0.941} & \uline{0.990} & \textbf{0.998}\tabularnewline
DeepV2D~\cite{Teed2020DeepV2DVT} & $300\times1088$ & \textbf{0.037} & \textbf{0.174} & \uline{2.005} & \textbf{0.074} & \textbf{0.977} & \textbf{0.993} & \uline{0.997}\tabularnewline
\hline 
PWCDC-Net~\cite{Sun2018PWCNetCF} & $256\times768$ & 0.152 & 2.015 & 5.883 & 0.251 & 0.828 & 0.920 & 0.956\tabularnewline
M4Depth-d6 (Ours) & $256\times768$ & 0.095 & 0.7084 & 3.515 & 0.146 & 0.898 & 0.962 & 0.982\tabularnewline
\hline 
\end{tabular}
\end{table*}
We observe that M4Depth has similar performances to current state-of-the-art
methods. As expected, instances with dynamic elements or poor GPS
localization lead to degraded performances. These results however
prove that M4Depth also works with real data despite their imperfections.

An overview of the outputs of our method on \kitti is shown in \Figx{\ref{fig:overview-kitti}}.
M4Depth appears to preserve fine details, and to estimated depth reliably
even in areas with less texture or for glossy objects such as cars.

\begin{figure}[!t]
\centering{}\setlength{\tabcolsep}{2pt}%
\begin{tabular}{cc}
\includegraphics[width=0.48\linewidth]{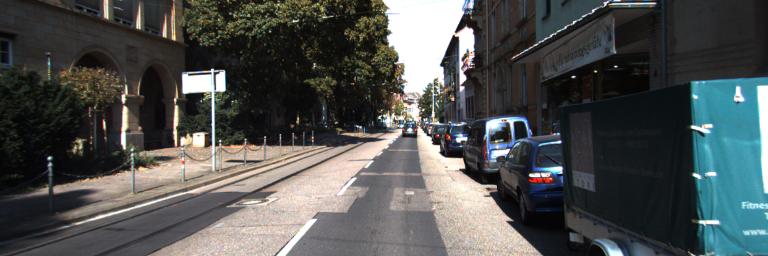} & \includegraphics[width=0.48\linewidth]{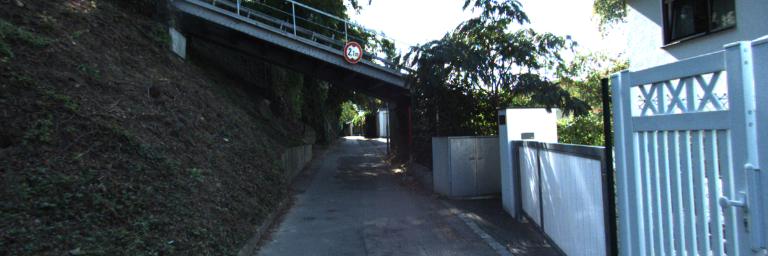}\tabularnewline
\includegraphics[width=0.48\linewidth]{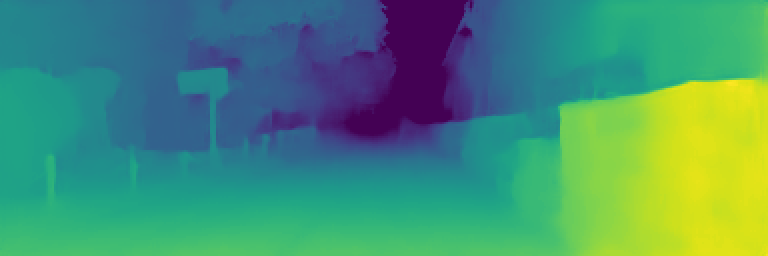} & \includegraphics[width=0.48\linewidth]{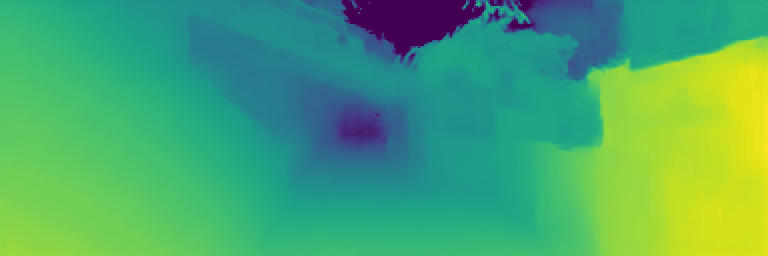}\tabularnewline
\includegraphics[width=0.48\linewidth]{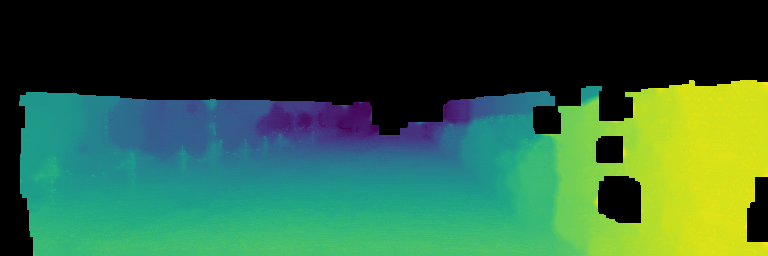} & \includegraphics[width=0.48\linewidth]{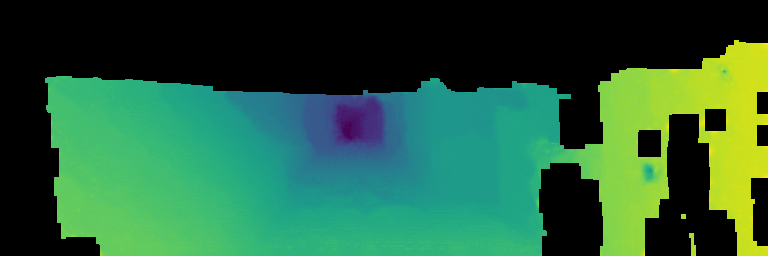}\tabularnewline
\end{tabular}\caption{Comparison of depth maps estimated by M4Depth on the \kitti dataset
(Row 2) with the corresponding interpolated ground truth (Row 3).\label{fig:overview-kitti}}
\end{figure}

\subsection{Generalization}

In this last experiment, we want to evaluate the generalization capabilities
of all the methods. For this, we want to use static scenes that are
semantically close to either the Mid-Air dataset (natural unstructured
environments) or the \kitti dataset (urban environments), and test
the performance of the method trained on \midair (respectively \kitti)
on the selected unstructured (respectively urban) scenes without any
fine-tuning.

\textbf{\tartanair dataset}~\cite{Wang2020TartanAir}\textbf{.}
For this experiment, we use \tartanair. It is a synthetic dataset
consisting of trajectories recorded by a free-flying camera in a series
of different environment scenes. With each scene being relatively
small in size, there is a lot of overlap in the visual content recorded
for different trajectories within a same scene. As such, assembling
clearly separated train and test sets drawn from the same data distribution
is not possible. Despite this drawback, the diversity of the scenes
makes \tartanair an interesting choice for testing the generalization
capabilities of methods.

For the generalization test from the Mid-Air dataset, we select the
``Gascola'' and ``Season forest winter'' scenes of \tartanair
and use the weights trained for the baseline. For the one from the
\kitti dataset, we select the ``Neighborhood'' and ``Old Town''
scenes and use the pre-trained weights released by the authors of
the methods.

We resized the images of this dataset to $384\times576$ pixels and
subsampled the frame rate by a factor of two. Additionally, some scenes
appeared to have large underexposed areas where there is no color
information in RGB frames. Having large pitch-black areas in an RGB
image is unrealistic in practice as cameras dynamically adapt their
shutter speed depending on the exposure of the scene. To prevent the
errors made by depth estimation methods in these areas from dominating
the performance analysis, we discarded all the pixels for which the
color in the RGB image with a value equal to zero.

\textbf{Results.} As we want to focus only on the generalization performance
for depth estimation, we bypass the pose estimation network for ManyDepth
and DeepV2D, and use the ground-truth motion to generate the depth
maps with these methods. Also, the depth maps produced by baseline
methods are not guaranteed to be at the correct scale. To alleviate
this issue in performance tests, we apply a median scaling to the
depth maps of baseline methods. The results of our experiments are
reported in \tabx{\ref{tab:perf_generalization}}. 
\begin{table*}[p]
\caption{Performance comparison in generalization on the TartanAir dataset.
\label{tab:perf_generalization}}
\subfloat[Performance for the generalization test on two unstructured environments
from \tartanair, that are gascola (G) and season forest winter (W).
Scores were generated by using the same network weights as the ones
used to report the performance on \midair in \tabx{\ref{tab:perf_scores-midair}}.\label{tab:perf_scores-generalization-unstruct}]{
\centering{}\begin{adjustbox}{width=0.99\linewidth}%
\begin{tabular}{|l|c||cc|cc|cc|cc||cc|cc|cc|}
\hline 
\multirow{2}{*}{Method} & \multirow{2}{*}{Test size} & \multicolumn{2}{c|}{\cellcolor{blue!25}Abs Rel $\downarrow$} & \multicolumn{2}{c|}{\cellcolor{blue!25}SQ Rel $\downarrow$} & \multicolumn{2}{c|}{\cellcolor{blue!25}RMSE $\downarrow$} & \multicolumn{2}{c||}{\cellcolor{blue!25}RMSE log $\downarrow$} & \multicolumn{2}{c|}{\cellcolor{orange!40}$\delta<1.25$ $\uparrow$} & \multicolumn{2}{c|}{\cellcolor{orange!40}$\delta<1.25^{2}$ $\uparrow$} & \multicolumn{2}{c|}{\cellcolor{orange!40}$\delta<1.25^{3}$ $\uparrow$}\tabularnewline
\cline{3-16} \cline{4-16} \cline{5-16} \cline{6-16} \cline{7-16} \cline{8-16} \cline{9-16} \cline{10-16} \cline{11-16} \cline{12-16} \cline{13-16} \cline{14-16} \cline{15-16} \cline{16-16} 
 &  & G & W & G & W & G & W & G & W & G & W & G & W & G & W\tabularnewline
\hline 
Monodepth~\cite{Godard2017UnsupervisedMD} & $384\times512$ & 0.929 & 1.765 & 21.950 & 147.3026 & 19.116 & 33.162 & 0.992 & 1.118 & 0.231 & 0.224 & 0.430 & 0.384 & 0.590 & 0.516\tabularnewline
Monodepth2~\cite{Godard2019DiggingIS} & $384\times384$ & 0.922 & 1.651 & 19.274 & 67.815 & 18.527 & 24.543 & 0.799 & 1.058 & 0.310 & 0.286 & 0.507 & 0.437 & 0.651 & 0.561\tabularnewline
ST-CLSTM~\cite{Zhang2019ExploitingTC} & $384\times384$ & 2.967 & 2.552 & 51.305 & 40.452 & 32.453 & 27.338 & 0.978 & \uline{0.878} & 0.375 & 0.370 & 0.517 & 0.518 & 0.626 & 0.609\tabularnewline
Wang~\cite{Wang2019RecurrentNN} & $384\times512$ & 0.761 & 0.776 & 25.459 & 29.138 & 31.875 & 28.332 & 1.482 & 1.260 & 0.209 & 0.259 & 0.313 & 0.353 & 0.411 & 0.442\tabularnewline
ManyDepth~\cite{Watson2021TheTemporal} & $384\times384$ & 0.776 & 1.383 & 16.551 & 63.285 & \uline{16.822} & \uline{23.607} & 0.746 & 0.974 & 0.326 & 0.374 & 0.538 & 0.544 & 0.684 & \uline{0.654}\tabularnewline
\hline 
PWCDC-Net~\cite{Sun2018PWCNetCF} & $384\times512$ & \uline{0.343} & \textbf{0.516} & \uline{7.645} & \uline{18.459} & 17.731 & 30.028 & \uline{0.684} & 1.160 & \uline{0.584} & \uline{0.463} & \uline{0.716} & \uline{0.568} & \uline{0.786} & 0.632\tabularnewline
M4Depth-d6 (Ours) & $384\times512$ & \textbf{0.281} & \uline{0.537} & \textbf{5.348} & \textbf{17.040} & \textbf{11.875} & \textbf{16.937} & \textbf{0.524} & \textbf{0.694} & \textbf{0.715} & \textbf{0.663} & \textbf{0.806} & \textbf{0.746} & \textbf{0.856} & \textbf{0.798}\tabularnewline
\hline 
\end{tabular}\end{adjustbox}}

\subfloat[Performance for the generalization test on two structured environments
from \tartanair, that are neighborhood (N) and old town (OT). Scores
were generated by using the same network weights as the ones used
to report the performance on \kitti in \tabx{\ref{tab:perf_scores-kitti}}.\label{tab:perf_scores-generalization-struct}]{
\centering{}\begin{adjustbox}{width=0.99\linewidth}%
\begin{tabular}{|l|c||cc|cc|cc|cc||cc|cc|cc|}
\hline 
\multirow{2}{*}{Method} & \multirow{2}{*}{Test size} & \multicolumn{2}{c|}{\cellcolor{blue!25}Abs Rel $\downarrow$} & \multicolumn{2}{c|}{\cellcolor{blue!25}SQ Rel $\downarrow$} & \multicolumn{2}{c|}{\cellcolor{blue!25}RMSE $\downarrow$} & \multicolumn{2}{c||}{\cellcolor{blue!25}RMSE log $\downarrow$} & \multicolumn{2}{c|}{\cellcolor{orange!40}$\delta<1.25$ $\uparrow$} & \multicolumn{2}{c|}{\cellcolor{orange!40}$\delta<1.25^{2}$ $\uparrow$} & \multicolumn{2}{c|}{\cellcolor{orange!40}$\delta<1.25^{3}$ $\uparrow$}\tabularnewline
\cline{3-16} \cline{4-16} \cline{5-16} \cline{6-16} \cline{7-16} \cline{8-16} \cline{9-16} \cline{10-16} \cline{11-16} \cline{12-16} \cline{13-16} \cline{14-16} \cline{15-16} \cline{16-16} 
 &  & N & OT & N & OT & N & OT & N & OT & N & OT & N & OT & N & OT\tabularnewline
\hline 
Monodepth~\cite{Godard2017UnsupervisedMD} & $384\times512$ & 1.041 & 0.909 & 54.683 & 30.616 & 30.957 & 19.203 & 0.843 & 0.782 & 0.261 & 0.267 & 0.465 & 0.489 & 0.620 & 0.655\tabularnewline
Monodepth2~\cite{Godard2019DiggingIS} & $320\times1024$ & 0.810 & 0.775 & 27.904 & 17.973 & 21.011 & 15.800 & \uline{0.732} & 0.727 & 0.412 & 0.322 & 0.603 & 0.540 & 0.715 & 0.692\tabularnewline
ManyDepth~\cite{Watson2021TheTemporal} & $320\times1024$ & 0.942 & 0.759 & 42.846 & 20.895 & 22.508 & 15.604 & 0.757 & 0.649 & 0.432 & 0.351 & 0.607 & 0.583 & 0.714 & 0.730\tabularnewline
DeepV2D~\cite{Teed2020DeepV2DVT} & $384\times512$ & 1.5157 & 0.694 & 77.063 & 18.777 & \uline{21.546} & \uline{7.551} & 0.769 & \uline{0.498} & 0.335 & 0.494 & 0.527 & 0.722 & 0.641 & \uline{0.830}\tabularnewline
\hline 
PWCDC-Net~\cite{Sun2018PWCNetCF} & $384\times512$ & \textbf{0.376} & \uline{0.338} & \textbf{12.66} & \uline{8.679} & 23.782 & 16.760 & 0.788 & 0.703 & \uline{0.535} & \uline{0.627} & \uline{0.652} & \uline{0.741} & \uline{0.723} & 0.800\tabularnewline
M4Depth (Ours) & $384\times512$ & \uline{0.509} & \textbf{0.256} & \uline{24.283} & \textbf{6.759} & \textbf{13.150} & \textbf{7.211} & \textbf{0.502} & \textbf{0.370} & \textbf{0.749} & \textbf{0.804} & \textbf{0.827} & \textbf{0.880} & \textbf{0.872} & \textbf{0.918}\tabularnewline
\hline 
\end{tabular}\end{adjustbox}}
\end{table*}
Overall M4Depth outperforms the other methods with a significant margin
both for structured and unstructured environments. As on \midair,
PWC-Net slightly outperforms M4Depth on some relative metrics, but
not for both sequences. It is worth noting that the hierarchy of the
performances has completely changed between the test on \kitti and
the one in generalization as our method outperforms DeepV2D~\cite{Teed2020DeepV2DVT}
on the latter. These results therefore show the better generalization
capability of M4Depth when compared to state-of-the-art methods.

\begin{table*}[!p]
\caption{Performance of M4Depth (trained on \midair, averaged over 4 runs)
for various architecture depths and ablations (on a network with 6
levels), and for a full architecture with $2$, $4$, and $6$ levels,
when tested on \midair (MA) as well as in generalization on the old
town scene (OT) of \tartanair.\label{tab:perf_ablation}}

\centering{}%
\begin{tabular}{|l||cc|cc|cc|cc||cc|cc|cc|}
\hline 
\multirow{2}{*}{Ablation} & \multicolumn{2}{c|}{\cellcolor{blue!25}Abs Rel $\downarrow$} & \multicolumn{2}{c|}{\cellcolor{blue!25}SQ Rel $\downarrow$} & \multicolumn{2}{c|}{\cellcolor{blue!25}RMSE $\downarrow$} & \multicolumn{2}{c||}{\cellcolor{blue!25}RMSE log $\downarrow$} & \multicolumn{2}{c|}{\cellcolor{orange!40}$\delta<1.25$ $\uparrow$} & \multicolumn{2}{c|}{\cellcolor{orange!40}$\delta<1.25^{2}$ $\uparrow$} & \multicolumn{2}{c|}{\cellcolor{orange!40}$\delta<1.25^{3}$ $\uparrow$}\tabularnewline
\cline{2-15} \cline{3-15} \cline{4-15} \cline{5-15} \cline{6-15} \cline{7-15} \cline{8-15} \cline{9-15} \cline{10-15} \cline{11-15} \cline{12-15} \cline{13-15} \cline{14-15} \cline{15-15} 
 & MA & OT & MA & OT & MA & OT & MA & OT & MA & OT & MA & OT & MA & OT\tabularnewline
\hline 
SNCV & 0.118 & 0.609 & 4.392 & 26.870 & 7.730 & 9.469 & 0.203 & 0.608 & 0.912 & 0.738 & 0.947 & 0.807 & 0.965 & 0.846\tabularnewline
Normalize & 0.099 & 0.583 & 3.179 & 23.495 & 7.032 & 9.019 & 0.185 & 0.496 & 0.920 & 0.770 & 0.955 & 0.842 & 0.971 & 0.880\tabularnewline
DINL & 0.104 & 0.521 & 3.480 & 19.378 & 7.182 & 8.826 & 0.189 & 0.536 & 0.915 & 0.763 & 0.952 & 0.833 & 0.969 & 0.872\tabularnewline
$\mathbf{f}_{\rho,t}^{l-1}$ & 0.113 & 0.435 & 4.007 & 14.445 & 7.382 & 8.732 & 0.196 & 0.517 & 0.916 & 0.771 & 0.950 & 0.840 & 0.967 & 0.878\tabularnewline
Split & 0.113 & 0.366 & 4.074 & 9.937 & 7.424 & 7.900 & 0.196 & 0.478 & 0.914 & 0.762 & 0.949 & 0.834 & 0.966 & 0.873\tabularnewline
$\dispmap_{t-1}^{l}$ & 0.107 & 0.435 & 3.482 & 15.071 & 7.201 & 8.836 & 0.197 & 0.437 & 0.911 & 0.788 & 0.949 & 0.863 & 0.966 & 0.900\tabularnewline
\hline 
M4Depth-d2 & 0.108 & 0.660 & 3.164 & 30.091 & 8.141 & 13.743 & 0.230 & 0.618 & 0.903 & 0.742 & 0.943 & 0.820 & 0.962 & 0.861\tabularnewline
M4Depth-d4 & 0.114 & 0.330 & 4.124 & 12.569 & 7.405 & 8.391 & 0.196 & 0.399 & 0.916 & 0.809 & 0.951 & 0.882 & 0.967 & 0.917\tabularnewline
M4Depth-d6 & 0.109 & 0.434 & 3.724 & 14.087 & 7.169 & 8.875 & 0.190 & 0.494 & 0.917 & 0.778 & 0.952 & 0.848 & 0.968 & 0.885\tabularnewline
\hline 
\end{tabular}
\end{table*}

\subsection{Discussion on the architecture}

\textbf{Ablation study.} We report the average performance over four
trainings for ablated versions of our architecture in \tabx{\ref{tab:perf_ablation}}.
The results show that the SNCV is the block that leads to the best
performance boost. This highlights the benefits of giving some spatial
information to the disparity refiners. The other blocks contribute
to improve either test or generalization performances, but not both
at the same time. As expected, the main contributors to generalization
performances are the DINL and the normalization layer.

Increasing the number of levels in the architecture improves the performances.
It should be noted, however, that the network tends to overfit the
training dataset, therefore leading to worse generalization performance
if the network gets too deep.

Overall, this ablation study shows that a compromise between performance
on the training dataset and performance in generalization has to be
made.

\textbf{Limitations.} With our approach, large areas with no repetitive
textures are prone to poor depth estimates. The feature matching performed
by our cost volumes matching can indeed become unstable if large areas
share exactly the same features. This can therefore lead to bad depth
estimates.

We mitigate this issue by using a multi-scale network and by including
an SNCV at each of its levels, but these solutions do not make our
network completely immune to this issue.

\textbf{Inference speed.} Our network has $4.5$~million parameters
and requires up to $500$~MB of GPU memory to run with six levels.
At inference time on Mid-Air, an NVidia Tesla V100 GPU needs $17$~ms
to process a single frame for a raw TensorFlow implementation. This
corresponds to $59$~frames per second which is roughly twenty-times
faster than DeepV2D, the best performing method on \kitti. Such inference
speed is compatible with the real-time constraints required for robotic
applications.

\textbf{Interpretation of the results. }As opposed to other methods,
our network is designed to exclusively use the relative difference
between feature vectors rather than relying on the raw semantic cues,
\ie, the raw value of the feature vectors, to estimate depth. All
reference methods, even the ones based on cost volumes, forward the
feature maps generated by their encoder directly to their depth estimation
subnetwork. Doing so gives networks the ability to use semantic cues
directly to estimate depth. This ability is only valuable for instances
where the set of features possibly encountered can be encoded by the
network and associated to a specific depth.

Our experiments show that reference methods perform well ---better
than M4Depth for some--- on \kitti, the dataset with constrained
and structured scenes. However, they fall behind in unstructured environments
when the link between semantic cues and depth is weak, and in generalization
when semantic cues are different from the reference. This tends to
imply that baseline networks rely on the raw feature values to derive
depth.

All these observations lead us to believe that severing the direct
link between the encoder and the decoder of the architecture while
proposing relevant substitute data through the preprocessing unit
is the key factor that allows M4Depth to perform so well overall in
our experiments.

\section{Conclusion\label{sec:Conclusion}}

In this paper, we address the challenging task of estimating depth
from RGB image sequences acquired in unknown environments by a camera
moving with six degrees of freedom. For this, we first define a notion
of visual disparity for generic camera motion, which is central for
our method M4Depth, and show how it can be used to estimate depth.
Then, we present new cost volumes designed to boost the performance
of the underlying deep neural network of our method.

Three series of experiments were performed on synthetic datasets as
well as on the \kitti dataset that features real data. They show
that M4Depth is superior to the baseline both in unstructured environments
and in generalization while also performing well on the standard \kitti
benchmark, which shows its superiority for autonomous vehicles that
would need to venture off road. In addition to being motion- and feature-invariant,
our method is lightweight, runs in real time, and can be trained in
an end-to-end fashion.

Our further works on M4Depth will, among others, focus on the determination
of its own uncertainty on depth estimates at inference time. Such
an addition would provide a great advantage over other methods that
do not offer this capability.

To help the community to reproduce our results, we made the code of
our method publicly available at \midairgithub.

\appendices{}

\ifcsname merge\endcsname
\begingroup
\let\clearpage\relax

\ifx \merge\False

\title{M4Depth: Monocular depth estimation for autonomous vehicles in unseen
environments ---~Supplementary material~---}
\author{Michaël Fonder, Damien Ernst and Marc Van Droogenbroeck}


\tableofcontents{}

\fi

\begingroup
\let\clearpage\relax

\ifx{ \charge\False \AND \merge\False}

\fi

\section{Geometry of the setup (complement to Section 4.1 Deriving depth from
visual disparity)}

\subsection{Camera model}

\label{sec:CameraModel}

In the pinhole camera model, the camera is represented by a sensor
plane and a focal point, which is the optical center of the camera,
taken as the origin (see \Figx{\ref{fig:pinhole}}). The focal point
is located somewhere along the principal axis that intersects the
sensor plane perpendicularly on its central point. The distance separating
the focal point from the sensor plane is the focal length; it is expressed
as a multiple of a sensor pixel width. 
\begin{figure}[ht]
\centering{}\includegraphics[width=0.4\textwidth]{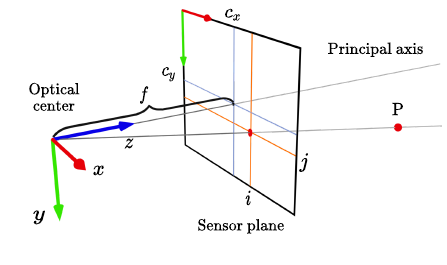}\caption{A diagram of the pinhole camera model with axes and notations. \label{fig:pinhole}}
\end{figure}

In its simple expression, the pinhole model of a camera is characterized
by five intrinsic parameters:
\begin{itemize}
\item $f_{x}$ and $f_{y}$, the focal lengths along the x and y axes, respectively;
\item $s$, the skew factor of a pixel, and
\item $(c_{x},c_{y})$, the coordinates of the principal point on the camera
sensor.
\end{itemize}
These parameters can be regrouped in a matrix $\mathbf{K}$, called
the \emph{projection} \emph{matrix}, as follows 
\begin{equation}
\mathbf{K}=\begin{bmatrix}f_{x} & s & c_{x}\\
0 & f_{y} & c_{y}\\
0 & 0 & 1
\end{bmatrix}\point
\end{equation}

The pixel coordinates $(i,j)$, in the camera plane, of the projection
of a point located at $(x,y,z)$ in the 3-D scene are obtained by
using the camera intrinsic matrix $\mathbf{K}$ and the right-angle
theorem as follows: 
\begin{equation}
(i,j)=\left(\frac{\alpha}{z},\frac{\beta}{z}\right)\:\text{with}\:\begin{bmatrix}\alpha\\
\beta\\
z\\
1
\end{bmatrix}=\begin{bmatrix}\mathbf{K} & \mathbf{0}\\
\mathbf{0} & 1
\end{bmatrix}\begin{bmatrix}x\\
y\\
z\\
1
\end{bmatrix}\point\label{equ:pinhole_proj}
\end{equation}
When the camera moves, the 3-D coordinates of a point can be expressed
not with respect to the current frame but with respect to the coordinates
system of the first frame. In this system, if we express the current
camera position by a vector $\mathbf{p}$ of size 3 and its orientation
by a $3\times3$ rotation matrix $\rotMatrix{}$, then, the position
of a point $(X,Y,Z)$ in space can be obtained by the following transformation
\begin{equation}
\begin{bmatrix}X\\
Y\\
Z\\
1
\end{bmatrix}=\underbrace{\begin{bmatrix}\rotMatrix{} & \mathbf{p}\\
\mathbf{0} & 1
\end{bmatrix}}_{\transformationMatrix{}}\begin{bmatrix}x\\
y\\
z\\
1
\end{bmatrix},
\end{equation}
where $\transpositionSymbol$ denotes the matrix transpose operator.
In this equation, $\transformationMatrix{}$ is called the \emph{transformation}
\emph{matrix}.

\subsection{Detailed derivation of the visual disparity}

\label{sec:Visual_disparity}

The definition of the (new) visual disparity notion involves two distinct
poses of the same camera. We denote these poses by $\text{C}_{1}$
and $\text{C}_{2}$. The $\text{C}_{2}$ pose is expressed relative
to the pose of $\text{C}_{1}$ and it is encoded by the transformation
matrix $\transformationMatrix 2$ with: 
\begin{equation}
\transformationMatrix 2=\begin{bmatrix}\rotMatrix{} & \mathbf{t}\\
\mathbf{0} & 1
\end{bmatrix}.
\end{equation}
\begin{figure}[t]
\centering{}\includegraphics[width=0.3\textwidth]{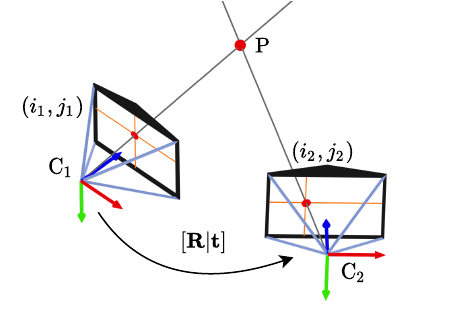}\caption{Illustration of moving the camera from position $C_{1}$ to $C_{2}$.\label{fig:triangulation}}
\end{figure}

Let us assume that some visual information is visible from two different
camera points of view, $\text{C}_{1}$ and $\text{C}_{2}$ (see \Figx{\ref{fig:triangulation}}).
A point $\text{P}$ seen by $\text{C}_{1}$ and projected on the sensor
plane in $(\hImCoord_{1},\vImCoord_{1})$ is projected at a different
location on the sensor of $\text{C}_{2}$, that is $(\hImCoord_{2},\vImCoord_{2})$.
For rigid camera motion, the camera intrinsic parameters are the same
for $\text{C}_{1}$ and $\text{C}_{2}$. Therefore, we simplify the
notations by expressing the coordinates $(\hImCoord,\vImCoord)$ relative
to the principal point $(c_{x},c_{y})$. With this definition and
by assuming that the skew parameter $s$ is negligible, which is common,
the expression of the camera intrinsic matrix $\mathbf{K}$ simplifies
into 
\begin{equation}
\mathbf{K}=\text{diag}(\focal x,\focal y,1)\,.
\end{equation}

To define our notion of visual disparity, we first need to find the
relation that links $(i_{1},j_{1})$ to $(\hImCoord_{2},\vImCoord_{2})$.
From \Eqx{\ref{equ:pinhole_proj}}, it can be seen that recovering
the full 3D coordinates of a point whose projection coordinates $(\hImCoord,\vImCoord)$
and depth $\dpix$ are known is trivial if the intrinsic matrix is
known (which is a common hypothesis in computer vision as long as
the camera is not zooming). Assuming that P is located at a depth
$\dpix_{2}$ of $\text{C}_{2}$, its 3D coordinates with respect to
$\text{C}_{2}$ are given by 
\begin{equation}
\mathbf{P}_{\text{C}_{2}}=\begin{bmatrix}\hImCoord_{2}/f_{x}\\
\vImCoord_{2}/f_{y}\\
1
\end{bmatrix}\dpix_{2}\,,\label{equ:reproj_1}
\end{equation}

These coordinates are expressed with respect to the $\text{C}_{2}$
referential. Their expression in $\text{C}_{1}$ is given by 
\begin{equation}
\mathbf{P}_{\text{C}_{1}}=[\rotMatrix{}|\translationVectorComp]\begin{bmatrix}\mathbf{P}_{\text{C}_{2}}\\
1
\end{bmatrix}=\rotMatrix{}\mathbf{P}_{\text{C}_{2}}+\translationVector\,.\label{equ:reproj_2}
\end{equation}
Computing the projection coordinates $(i_{1},j_{1})$ in $\text{C}_{1}$
for P is obtained by applying the camera projection equation (\Eqx{\ref{equ:pinhole_proj}}),
that is 
\begin{equation}
\dpix_{1}\begin{bmatrix}i_{1}\\
j_{1}\\
1
\end{bmatrix}=\mathbf{K}\mathbf{P}_{\text{C}_{1}}.\label{equ:reproj_3}
\end{equation}
By combining \Eqxyz{\ref{equ:reproj_1}}{\ref{equ:reproj_2}}{\ref{equ:reproj_3}},
we have

\begin{equation}
\dpix_{1}\begin{bmatrix}i_{1}\\
j_{1}\\
1
\end{bmatrix}=\mathbf{K}\left(\rotMatrix{}\ \dpix_{2}\begin{bmatrix}i_{2}/f_{x}\\
j_{2}/f_{y}\\
1
\end{bmatrix}+\translationVector\right)\,.\label{equ:reproj_tot}
\end{equation}
Equation~\ref{equ:reproj_tot} gives us the relationship between
the two coordinates $(\hImCoord_{1},\vImCoord_{1})$ and $(\hImCoord_{2},\vImCoord_{2})$
for a given location in the 3D space. This can be rewritten by introducing
an intermediate point $(i_{\indexOfVirtualCamera},j_{\indexOfVirtualCamera})$,
as

\begin{equation}
\begin{aligned}\dpix_{1}\begin{bmatrix}\hImCoord_{1}\\
\vImCoord_{1}\\
1
\end{bmatrix}= & \ \dpix_{2}\ \mathbf{K}\ \rotMatrix{}\begin{bmatrix}\hImCoord_{2}/\focal x\\
\vImCoord_{2}/\focal y\\
1
\end{bmatrix}+\mathbf{K}\ \begin{bmatrix}\translationVectorComp_{x}\\
\translationVectorComp_{y}\\
\translationVectorComp_{z}
\end{bmatrix}\\
= & \ \dpix_{2}\ \dpix_{R}\begin{bmatrix}\hImCoord_{\indexOfVirtualCamera}\\
\vImCoord_{\indexOfVirtualCamera}\\
1
\end{bmatrix}+\begin{bmatrix}\focal x\translationVectorComp_{x}\\
\focal y\translationVectorComp_{y}\\
\translationVectorComp_{z}
\end{bmatrix}\,,
\end{aligned}
\label{equ:p2p-cor}
\end{equation}
with 
\begin{equation}
\begin{bmatrix}\dpix_{\indexOfVirtualCamera}\hImCoord_{\indexOfVirtualCamera}\\
\dpix_{\indexOfVirtualCamera}\vImCoord_{\indexOfVirtualCamera}\\
\dpix_{\indexOfVirtualCamera}
\end{bmatrix}=\mathbf{K}\ \rotMatrix{}\ \begin{bmatrix}\hImCoord_{2}/\focal x\\
\vImCoord_{2}/\focal y\\
1
\end{bmatrix}\,.
\end{equation}
From this equation, we see that $(i_{\indexOfVirtualCamera},j_{\indexOfVirtualCamera})$
are the coordinates of point $P$ in the plane of a virtual camera
$\text{C}_{\indexOfVirtualCamera}$ whose origin is the same as camera
$\text{C}_{2}$ but with the orientation of camera $\text{C}_{1}$.
The plane of the virtual camera is thus parallel to that of camera
1. We now define our disparity $\disp$ as the Euclidean norm 
\begin{equation}
\disp_{ij}=\sqrt{\Delta_{i}^{2}+\Delta_{j}^{2}}\,,\label{equ:disp_app}
\end{equation}
where 
\begin{equation}
\left[\begin{array}{c}
\Delta_{i}\\
\Delta_{j}
\end{array}\right]=\left[\begin{array}{c}
i_{1}-i_{\indexOfVirtualCamera}\\
j_{1}-j_{\indexOfVirtualCamera}
\end{array}\right]\,.\label{equ:flow_app}
\end{equation}

We insert the expression of \Eqx{\ref{equ:flow_app}} in \Eqx{\ref{equ:p2p-cor}},
that is 
\begin{equation}
\dpix_{1}\begin{bmatrix}\hImCoord_{\indexOfVirtualCamera}+\Delta_{i}\\
\vImCoord_{\indexOfVirtualCamera}+\Delta_{j}\\
1
\end{bmatrix}=\ \dpix_{2}\ \dpix_{\indexOfVirtualCamera}\begin{bmatrix}\hImCoord_{\indexOfVirtualCamera}\\
\vImCoord_{\indexOfVirtualCamera}\\
1
\end{bmatrix}+\begin{bmatrix}\focal x\translationVectorComp_{x}\\
\focal y\translationVectorComp_{y}\\
\translationVectorComp_{z}
\end{bmatrix}.\label{equ:p2p-flow}
\end{equation}
It follows that 
\begin{equation}
\dpix_{1}=\dpix_{2}\ \dpix_{\indexOfVirtualCamera}+\translationVectorComp_{z}\comma
\end{equation}
and 
\begin{equation}
\begin{bmatrix}\Delta_{i}\\
\Delta_{j}
\end{bmatrix}=\frac{1}{\dpix_{2}\ \dpix_{\indexOfVirtualCamera}+\translationVectorComp_{z}}\begin{bmatrix}\focal x\translationVectorComp_{x}-\translationVectorComp_{z}\hImCoord_{\indexOfVirtualCamera}\\
\focal y\translationVectorComp_{y}-\translationVectorComp_{z}\vImCoord_{\indexOfVirtualCamera}
\end{bmatrix}.\label{equ:flow-1}
\end{equation}
The disparity $\disp_{ij}$ then becomes 
\begin{equation}
\begin{aligned}\disp_{ij}= & \frac{\sqrt{\left(\focal x\translationVectorComp_{x}-\translationVectorComp_{z}\hImCoord_{\indexOfVirtualCamera}\right)^{2}+\left(\focal y\translationVectorComp_{y}-\translationVectorComp_{z}\vImCoord_{\indexOfVirtualCamera}\right)^{2}}}{\left|\dpix_{2}\ \dpix_{\indexOfVirtualCamera}+\translationVectorComp_{z}\right|}\,.\end{aligned}
\label{equ:disp2depth_app}
\end{equation}

It is calculable for any point $P$ whose projection coordinates $(\hImCoord_{2},\vImCoord_{2})$
are within the image frame of $\text{C}_{2}$ and when the camera
motion is known, which is our setup. The only difficulty is the indetermination
of the sign of $\dpix_{2}\ \dpix_{\indexOfVirtualCamera}+\translationVectorComp_{z}$.
Theoretically, $\dpix_{2}\ \dpix_{\indexOfVirtualCamera}+\translationVectorComp_{z}$
can be negative, which is achievable, physically,  only if point
$P$ is located between the camera planes of $\text{C}_{1}$ and $\text{C}_{\indexOfVirtualCamera}$.
This situation is highly specific in practice, and should rarely occur
in the context of autonomous vehicles applications because it is never
advisable to move backwards, nor to move close to an object. From
the implementation point of view of our method, since such a point
$P$ is not visible to $\text{C}_{1}$, it is discarded during the
calculation of the cost volumes because it does not have any correspondence
in $\text{C}_{1}$. As a result, we use the following variant of \Eqx{\ref{equ:disp2depth_app}}
in practice

\begin{equation}
\disp_{ij}=\frac{\sqrt{\left(\focal x\translationVectorComp_{x}-\translationVectorComp_{z}\hImCoord_{\indexOfVirtualCamera}\right)^{2}+\left(\focal y\translationVectorComp_{y}-\translationVectorComp_{z}\vImCoord_{\indexOfVirtualCamera}\right)^{2}}}{\dpix_{2}\ \dpix_{\indexOfVirtualCamera}+\translationVectorComp_{z}}\,.\label{equ:disp2depth_approx}
\end{equation}

This last equation is useful because it links the disparity that should
be observed for a pixel and the depth of the corresponding point in
space. Since disparity can be observed and estimated from the two
images, we have a mean to estimate the depth by inverting the equation,
which yields 
\begin{equation}
\dpix_{2}=\frac{\sqrt{\left(\focal x\translationVectorComp_{x}-t_{z}\hImCoord_{\indexOfVirtualCamera}\right)^{2}+\left(\focal y\translationVectorComp_{y}-\translationVectorComp_{z}\vImCoord_{\indexOfVirtualCamera}\right)^{2}}}{\disp_{ij}\ \dpix_{\indexOfVirtualCamera}}-\frac{\translationVectorComp_{z}}{\dpix_{\indexOfVirtualCamera}}\,.\label{equ:depth2disp_app}
\end{equation}

In summary, \Eqxy{\ref{equ:disp2depth_app}}{\ref{equ:depth2disp_app}}
give the relationships to convert a depth map to a disparity map and
vice-versa for given pixel sensor coordinates and camera motion. These
relations are valid only if point $\mathbf{P}$ is static. In practice,
in our architecture, we respectively use the notations $\text{C}_{t-1}$
and $\text{C}_{t}$ instead of $\text{C}_{1}$ and $\text{C}_{2}$.

\endgroup

\section{Additional results (complement to Section 5 Experiments)}

\label{sec:Additional_overview}

In this section, we provide an additional overview and analysis on
the depth map estimates produced by our network with six levels.

\subsection{\kitti}

In \Figx{\ref{fig:overview-kitti}}, we show several outputs produced
by M4Depth on the test set of \kitti when trained on \midair and
fine-tuned on \kitti. Despite being trained in a supervised fashion
on sparse data, our network manages to make accurate predictions on
the whole image. 
\begin{figure*}[!tp]
\centering{}\setlength{\tabcolsep}{2pt}%
\begin{tabular}{cccc}
\includegraphics[width=0.24\linewidth]{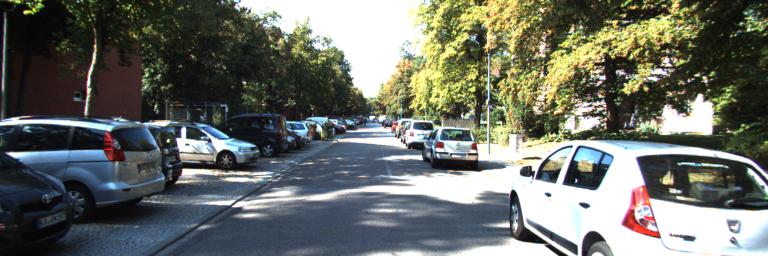} & \includegraphics[width=0.24\linewidth]{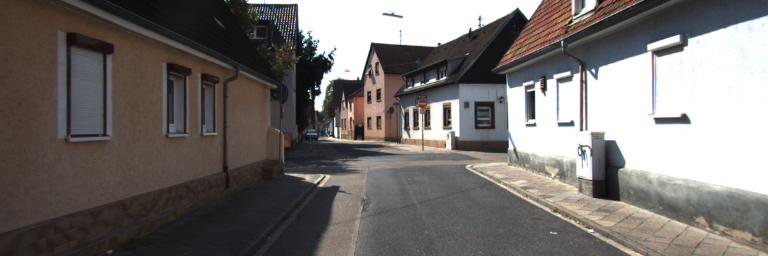} & \includegraphics[width=0.24\linewidth]{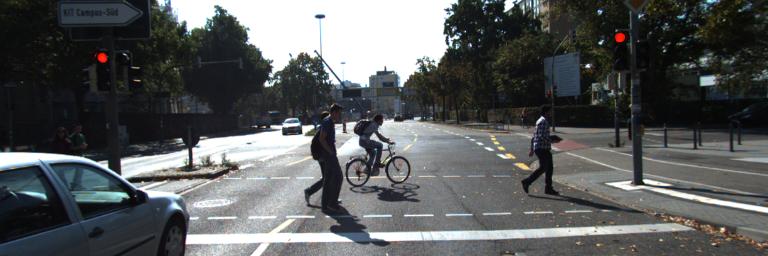} & \includegraphics[width=0.24\linewidth]{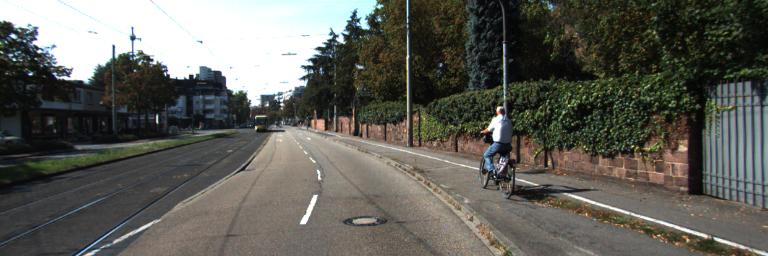}\tabularnewline
\includegraphics[width=0.24\linewidth]{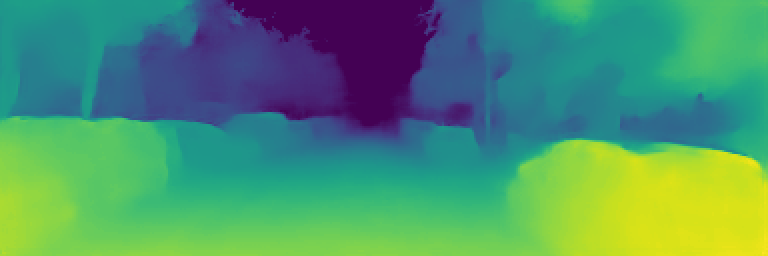} & \includegraphics[width=0.24\linewidth]{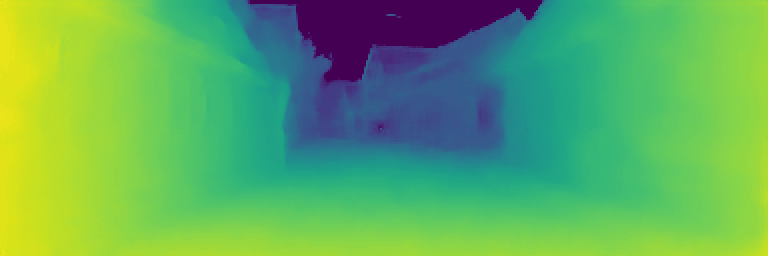} & \includegraphics[width=0.24\linewidth]{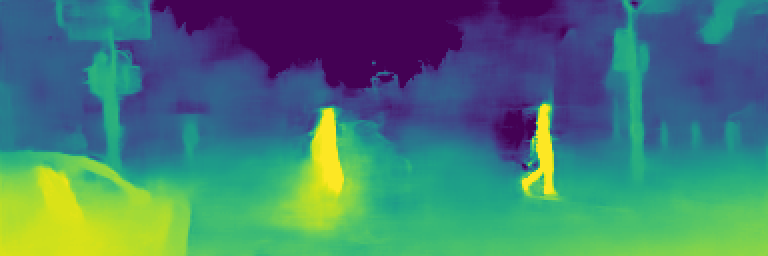} & \includegraphics[width=0.24\linewidth]{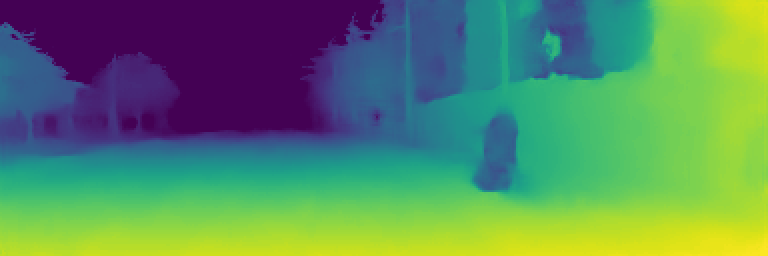}\tabularnewline
\includegraphics[width=0.24\linewidth]{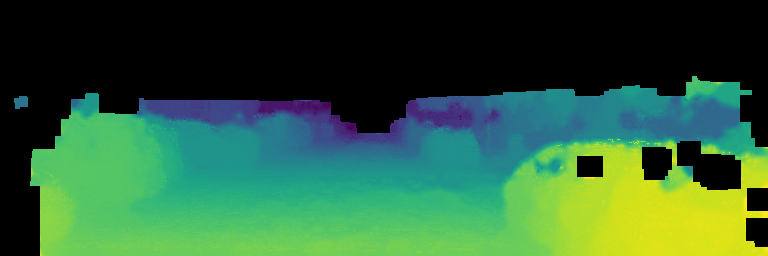} & \includegraphics[width=0.24\linewidth]{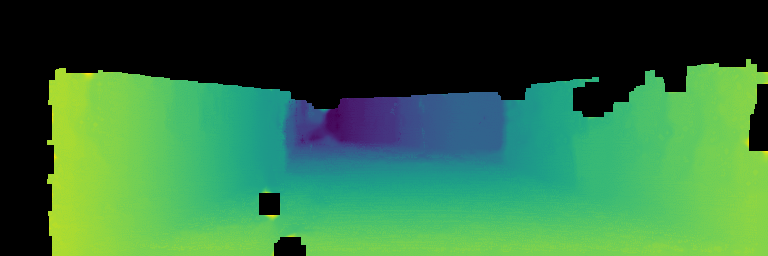} & \includegraphics[width=0.24\linewidth]{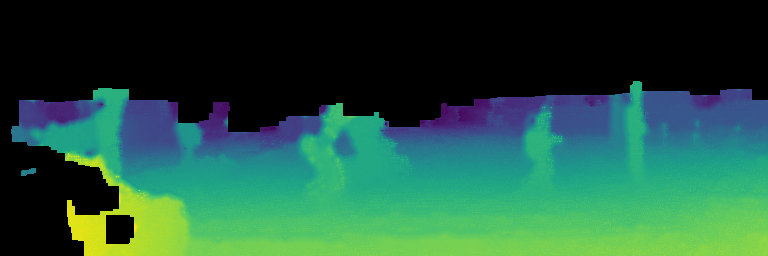} & \includegraphics[width=0.24\linewidth]{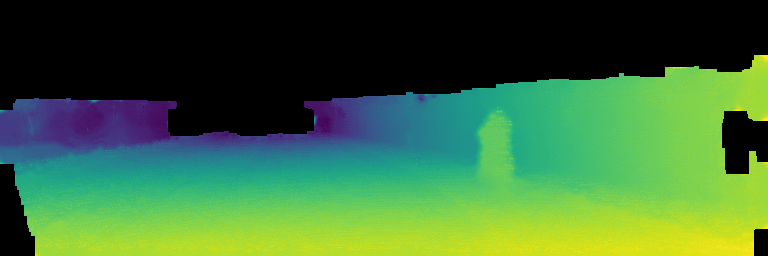}\tabularnewline
\end{tabular}\caption{Comparison of depth estimates produced by M4Depth on the \kitti test
set (Row 2) with the corresponding interpolated ground truth (Row
3). The two rightmost samples show artifacts produced by the network
around elements of the scene that are dynamic. \label{fig:overview-kitti}}
\end{figure*}

The first column shows that the network can deal with reflective surfaces,
and the second that the depth of textureless areas can also be correctly
estimated. However, these surfaces remain challenging for the network
and are not always handled as well as shown in these examples.

In our problem statement, we made the hypothesis that environments
are static. The two rightmost samples show the behavior of M4Depth
when this hypothesis is not met. The depth estimated for mobile objects
is either largely under- or overestimated depending on the relative
motion perceived by the camera. As we use a pyramidal architecture,
the reduction of the spatial dimension in the deeper layers of the
architecture can lead to depth estimation artifacts that bleed around
mobile objects, as seen in the third column.

\subsection{Generalization}

Some samples of our generalization experiments on unstructured and
urban scenes of the \tartanair dataset are shown in \Figxy{\ref{fig:overview-tta-unstruct}}{\ref{fig:overview-tta-urban}}
respectively. 
\begin{figure*}[!tp]
\subfloat[Gascola scene.]{\begin{adjustbox}{width=\columnwidth}%
\begin{minipage}[t]{1.1\columnwidth}%
\setlength{\tabcolsep}{2pt}%
\begin{tabular}{ccc}
\includegraphics[width=0.32\textwidth]{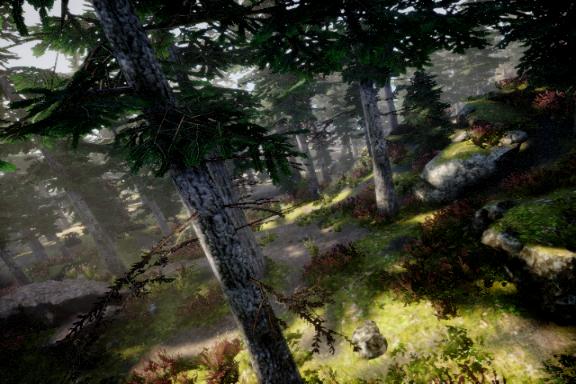} & \includegraphics[width=0.32\textwidth]{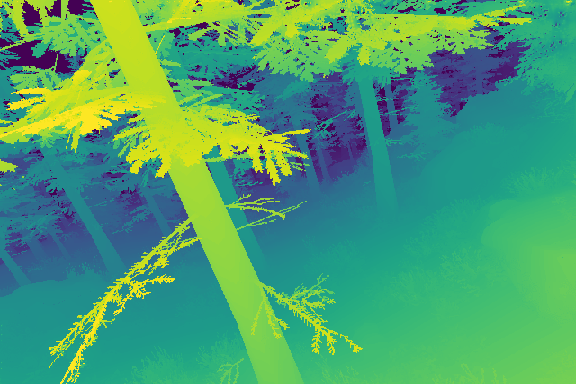} & \includegraphics[width=0.32\textwidth]{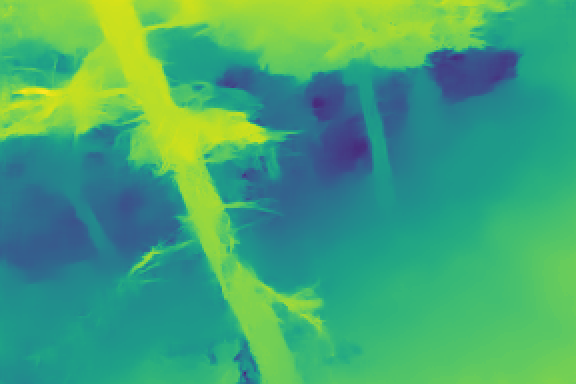}\tabularnewline
\includegraphics[width=0.32\textwidth]{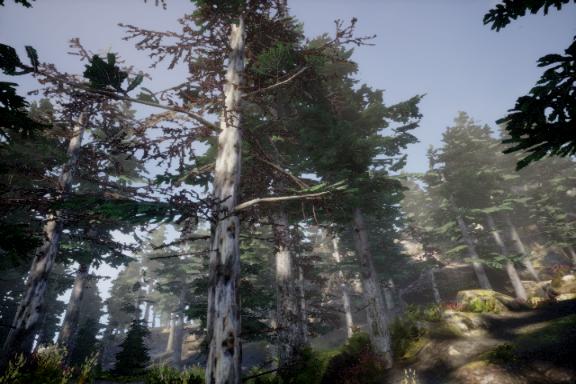} & \includegraphics[width=0.32\textwidth]{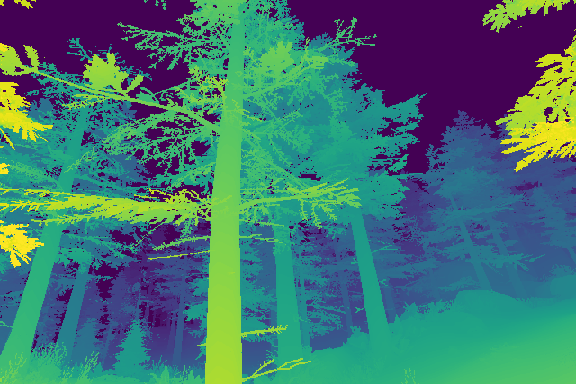} & \includegraphics[width=0.32\textwidth]{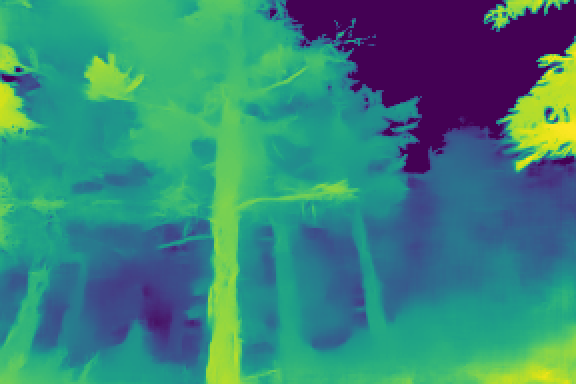}\tabularnewline
\includegraphics[width=0.32\textwidth]{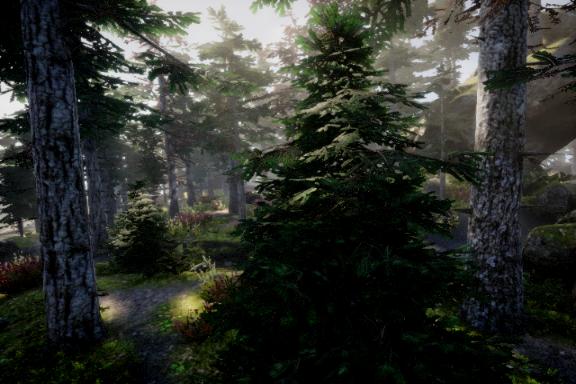} & \includegraphics[width=0.32\textwidth]{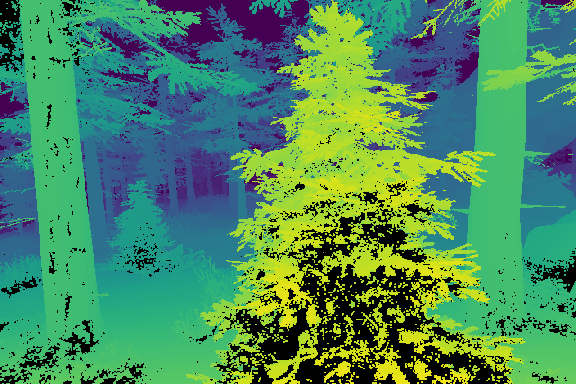} & \includegraphics[width=0.32\textwidth]{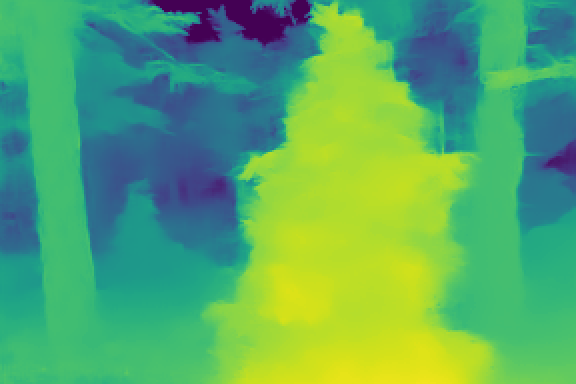}\tabularnewline
\vspace{-0.3cm}
 &  & \tabularnewline
RGB image & Ground truth & M4Depth\tabularnewline
\end{tabular}%
\end{minipage}\end{adjustbox}}\hfill{}\subfloat[Season forest (winter) scene.]{\begin{adjustbox}{width=\columnwidth}%
\begin{minipage}[t]{1.1\columnwidth}%
\setlength{\tabcolsep}{2pt}%
\begin{tabular}{ccc}
\includegraphics[width=0.32\textwidth]{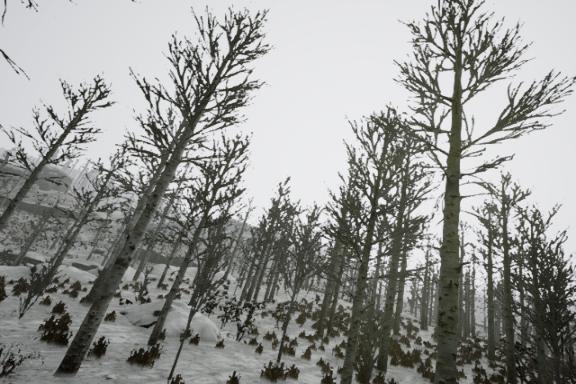} & \includegraphics[width=0.32\textwidth]{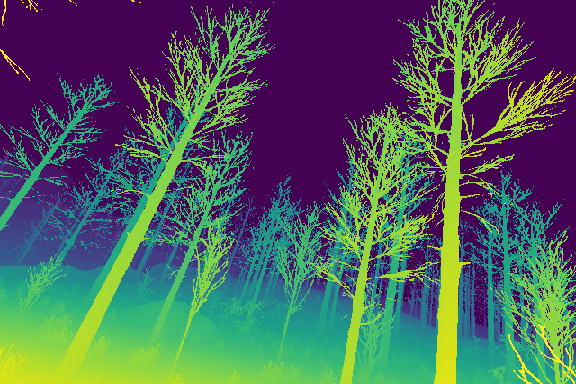} & \includegraphics[width=0.32\textwidth]{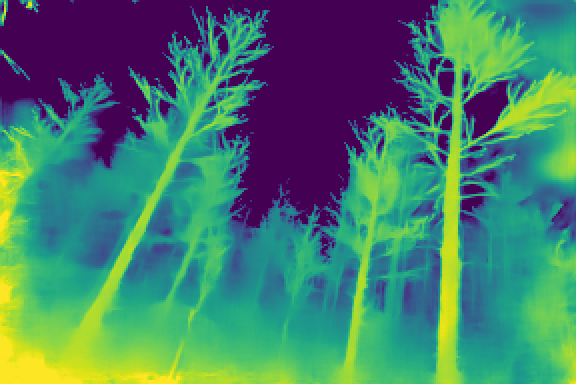}\tabularnewline
\includegraphics[width=0.32\textwidth]{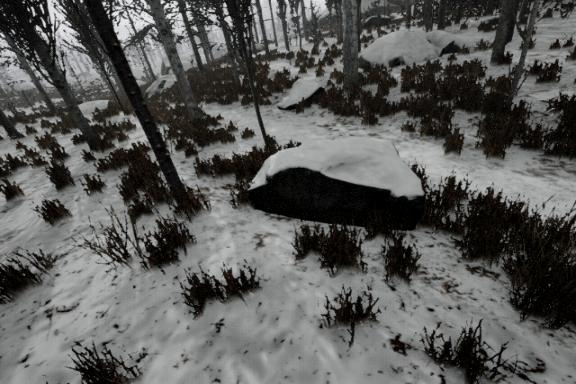} & \includegraphics[width=0.32\textwidth]{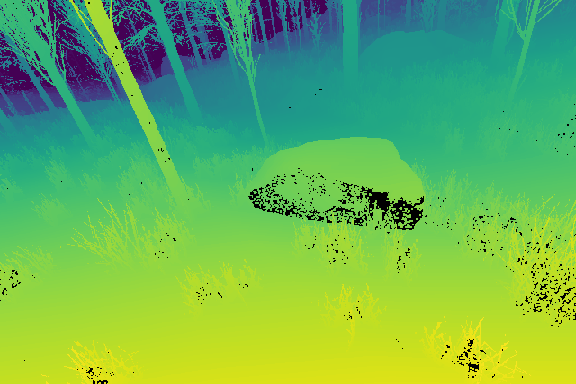} & \includegraphics[width=0.32\textwidth]{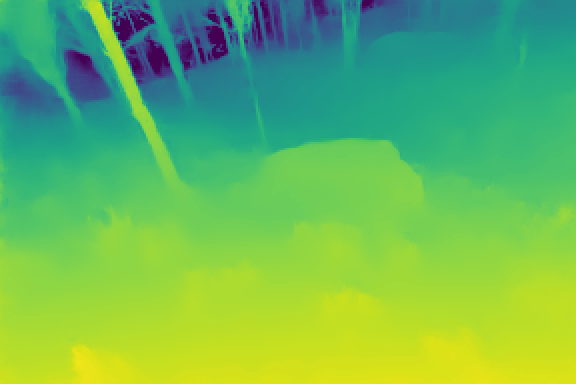}\tabularnewline
\includegraphics[width=0.32\textwidth]{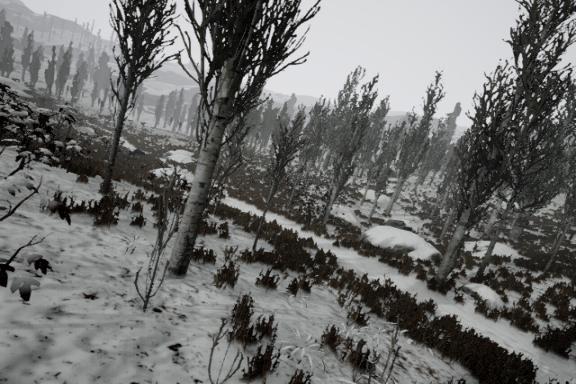} & \includegraphics[width=0.32\textwidth]{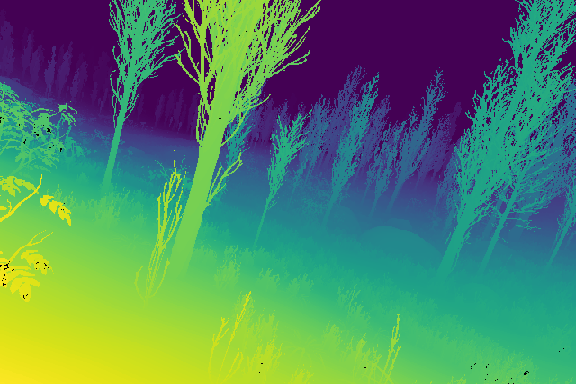} & \includegraphics[width=0.32\textwidth]{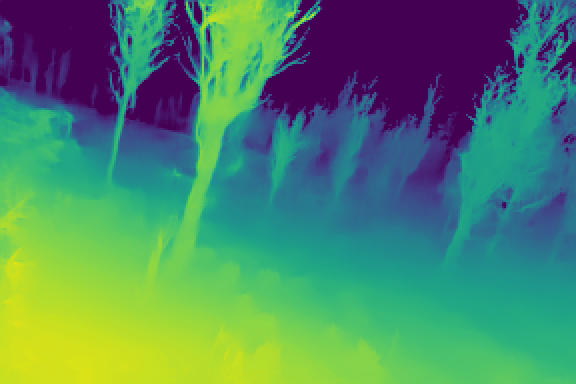}\tabularnewline
\vspace{-0.3cm}
 &  & \tabularnewline
RGB image & Ground truth & M4Depth\tabularnewline
\end{tabular}%
\end{minipage}\end{adjustbox}}

\caption{Samples of depth maps produced in generalization on unstructured scenes
of the \tartanair dataset by M4Depth with six levels trained on \midair.
Black areas in the ground truths correspond to pixels with no color
information in the RGB image.\label{fig:overview-tta-unstruct}}
\end{figure*}
\begin{figure*}[!tp]
\subfloat[Neighborhood scene.]{\begin{adjustbox}{width=\columnwidth}%
\begin{minipage}[t]{1.1\columnwidth}%
\setlength{\tabcolsep}{2pt}%
\begin{tabular}{ccc}
\includegraphics[width=0.32\textwidth]{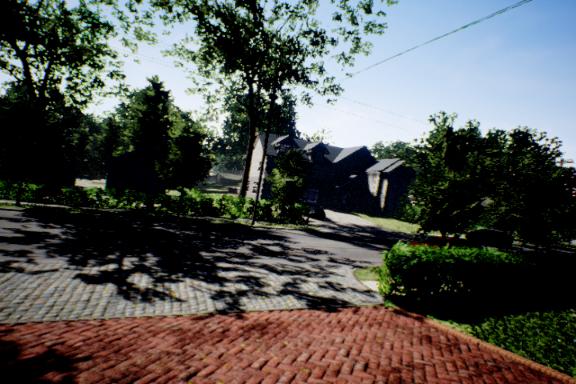} & \includegraphics[width=0.32\textwidth]{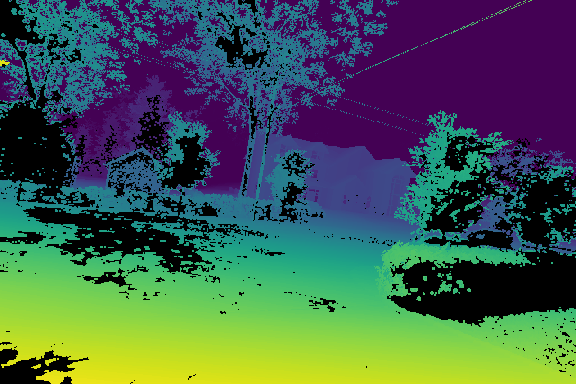} & \includegraphics[width=0.32\textwidth]{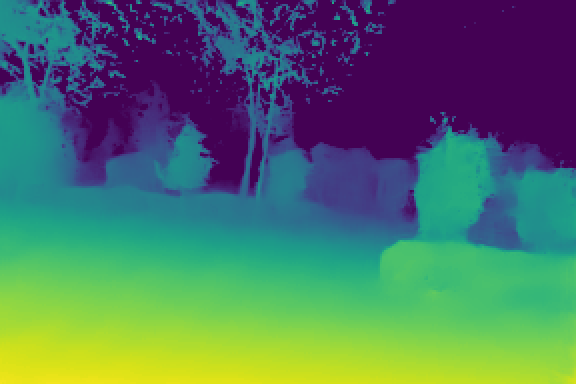}\tabularnewline
\includegraphics[width=0.32\textwidth]{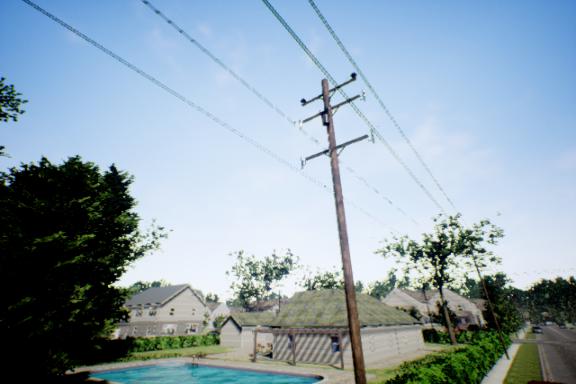} & \includegraphics[width=0.32\textwidth]{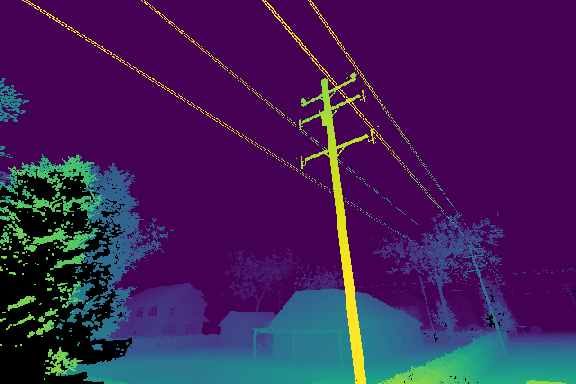} & \includegraphics[width=0.32\textwidth]{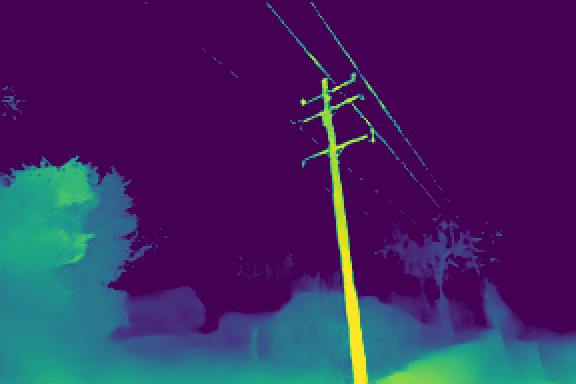}\tabularnewline
\includegraphics[width=0.32\textwidth]{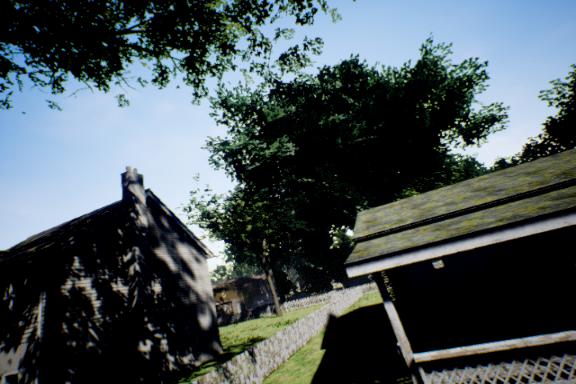} & \includegraphics[width=0.32\textwidth]{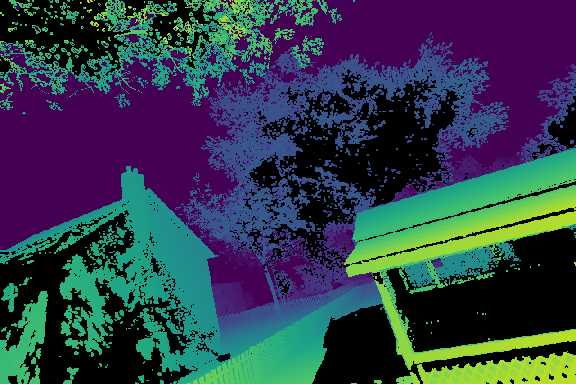} & \includegraphics[width=0.32\textwidth]{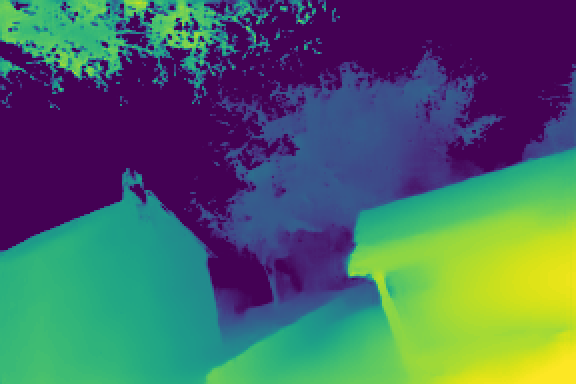}\tabularnewline
\vspace{-0.3cm}
 &  & \tabularnewline
RGB image & Ground truth & M4Depth\tabularnewline
\end{tabular}%
\end{minipage}\end{adjustbox}}\hfill{}\subfloat[Old town scene.]{\begin{adjustbox}{width=\columnwidth}%
\begin{minipage}[t]{1.1\columnwidth}%
\setlength{\tabcolsep}{2pt}%
\begin{tabular}{ccc}
\includegraphics[width=0.32\textwidth]{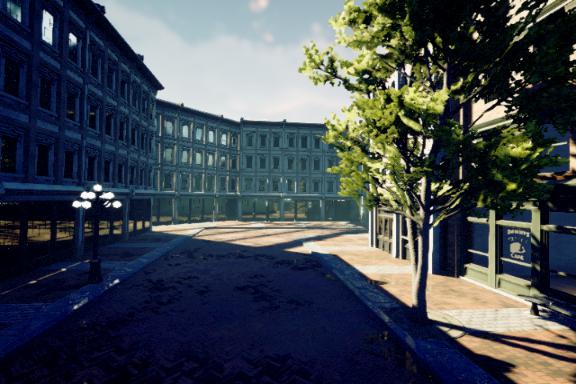} & \includegraphics[width=0.32\textwidth]{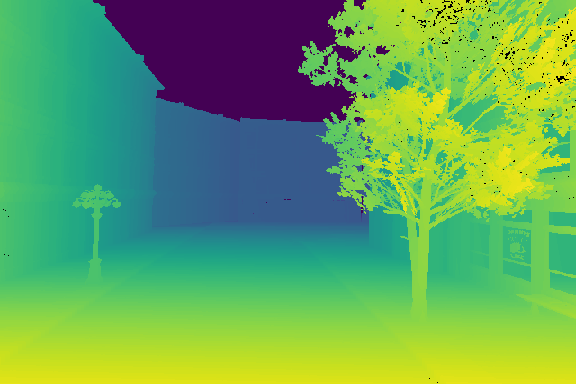} & \includegraphics[width=0.32\textwidth]{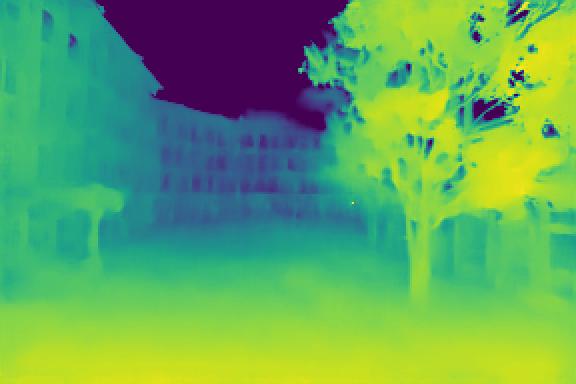}\tabularnewline
\includegraphics[width=0.32\textwidth]{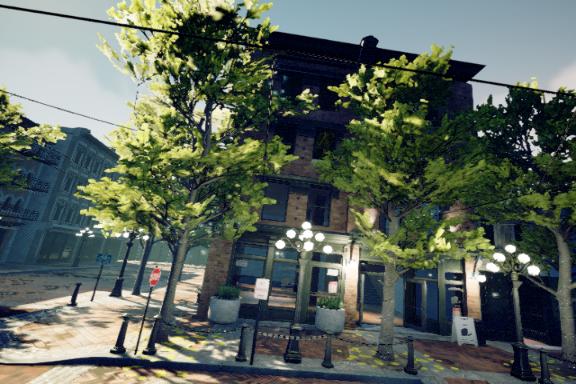} & \includegraphics[width=0.32\textwidth]{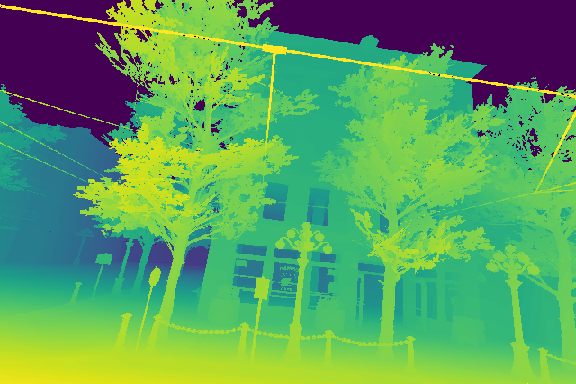} & \includegraphics[width=0.32\textwidth]{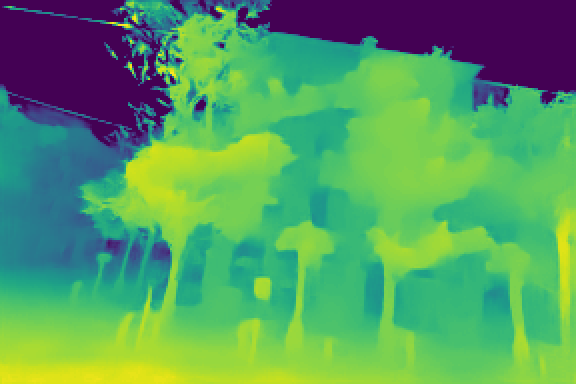}\tabularnewline
\includegraphics[width=0.32\textwidth]{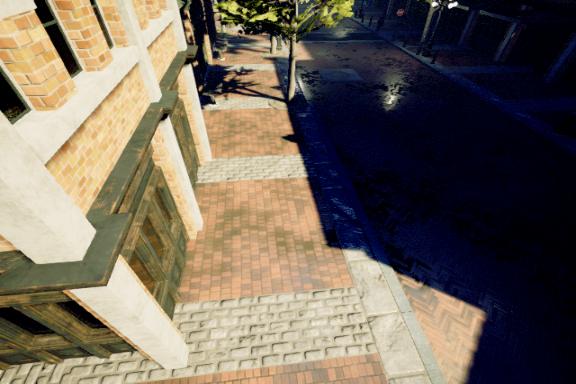} & \includegraphics[width=0.32\textwidth]{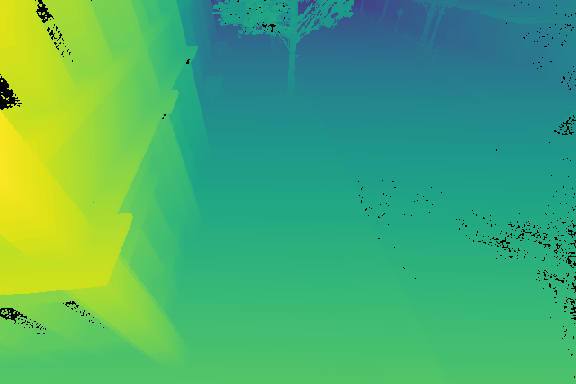} & \includegraphics[width=0.32\textwidth]{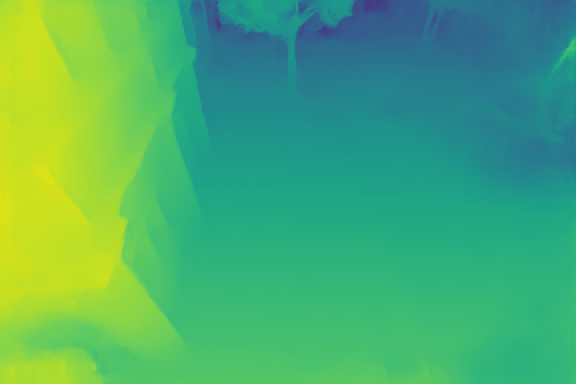}\tabularnewline
\vspace{-0.3cm}
 &  & \tabularnewline
RGB image & Ground truth & M4Depth\tabularnewline
\end{tabular}%
\end{minipage}\end{adjustbox}}

\caption{Samples of depth maps produced in generalization on urban scenes of
the \tartanair dataset by M4Depth with six levels trained on \midair
and fine-tuned on\foreignlanguage{english}{ }\kitti. Black areas
in the ground truths correspond to pixels with no color information
in the RGB image.\label{fig:overview-tta-urban}}
\end{figure*}

The quality of these results confirms the good scores obtained on
performance metrics. Objects in the scene are properly identified,
correctly detoured, and their depth is accurately estimated.

Some general observations on the weaknesses of M4Depth can also be
made from these outputs. First, the network cannot resolve all the
details when the scene is too cluttered. This is especially visible
in forest environments where tree branches overlap. Second, there
are sometimes issues with sky recognition. This is however to be expected
as our network mostly relies on perceived frame-to-frame pixel displacement
to produce estimates. Finally, small and isolated structures such
as cables are not always detected (see the outputs on urban scenes).

\section{Mid-Air baseline methods training details (complement to Section
5.1 Unstructured environments)}

In this section, we provide the training details to reproduce the
results of all the methods of our baseline for the \midair dataset.
These details complement the code made available on a GitHub repository.
In the paper, we have chosen five methods for our baseline, namely:
Monodepth~\cite{Godard2017UnsupervisedMD,Godard2017Monodepth}, Monodepth2~\cite{Godard2019DiggingIS,Godard2019Monodepth2},
ST-CLSTM~\cite{Zhang2019ExploitingTC,Weihaox2020}, the method of
Wang \etal~\cite{Wang2019RecurrentNN,Wang2019RNN}, ManyDepth~\cite{Watson2021TheTemporal,Watson2021ManyDepth},
and PWCDC-Net~\cite{Sun2018PWCNetCF}.

To get a baseline that is true to the work of the authors and is coherent
between different methods, we proceeded as follows. We kept the original
default parameter values of each method. Next, we adjusted the batch
size so that every learning step contained around $18$ frames, and
we trained each network five times. As some method have specific input
pipelines, we adjusted the training epoch count of each method to
guarantee that a network sees every training sample at least $50$
times during its training. After a first round of training, it appeared
that some methods did not converge. This has led us to adapt the training
setup for these methods in order to obtain a representative performance.

The training of PWCDC-Net~\cite{Sun2018PWCNetCF} had to be done
differently as it is an optical flow network. To make it work, we
had to convert depth maps to optical flow maps by using \Eqx{\ref{equ:reproj_tot}}.
It also required more steps during the training to reach a steady-state
on the validation set.

\begin{table}[t]
\caption{Main parameters used to train baseline methods. The asterisk denotes
default parameters of methods suggested by their respective authors.\label{tab:baseline_settings}}

\centering{}\begin{adjustbox}{width=0.99\linewidth}%
\begin{tabular}{|l|c|c|c|}
\hline 
Method & Train epoch count & Batch size & Sequence length\tabularnewline
\hline 
Monodepth~\cite{Godard2017UnsupervisedMD,Godard2017Monodepth} & 50 & 18 & 1\**\tabularnewline
Monodepth2~\cite{Godard2019DiggingIS,Godard2019Monodepth2} & 17 & 6 & 3\**\tabularnewline
ST-CLSTM~\cite{Zhang2019ExploitingTC,Weihaox2020} & 50 & 3 & 5\**\tabularnewline
Wang \etal~\cite{Wang2019RecurrentNN,Wang2019RNN} & 50 & 3 & 8\textcolor{white}{\**}\tabularnewline
Manydepth~\cite{Watson2021TheTemporal,Watson2021ManyDepth} & 25 & 6 & 3\**\tabularnewline
\hline 
PWCDC-Net~\cite{Sun2018PWCNetCF} & 100 & 8 & 2\**\tabularnewline
\hline 
\end{tabular}\end{adjustbox}
\end{table}

The values reported for the baseline methods correspond to the best
results obtained out of five runs. The most important parameters of
the baseline setup are given in Table~\ref{tab:baseline_settings}.
We kept all other parameters unchanged to a large extent. However,
adjustments were necessary for some methods as explained hereafter.
\begin{itemize}
\item With the proposed setup, Monodepth failed to produce any output for
three trainings out of the five.
\item The lower epoch count for Monodepth2 is due to the fact that the training
pipeline sees each sample three times during a single epoch.
\item Performances obtained with the default learning rate for the ST-CLSTM
method were extremely poor. We obtained better results by reducing
it to $10^{-4}$.
\item The code written by Wang~ \etal worked as expected. However, the
length of the training sequence had to be downsized from 10 to 8 frames
to accommodate our internal pipeline constraints.
\item Finally, for Manydepth, we had to select the encoder architecture;
we chose the ResNet-50 encoder.
\end{itemize}
\ifx \merge\False 

\bibliographystyle{IEEEtran}
\bibliography{bib/abbreviation,abbreviation-paper,bib/dataset,bib/depth,bib/drone,bib/labo,bib/learning,bib/optical-flow,bib/robotics,bib/segmentation,bib/vision}

\fi

\endgroup
\fi

\bibliographystyle{IEEEtran}

\end{document}